\documentclass[journal]{IEEEtran}
\usepackage{amsmath,amssymb}
\usepackage{graphicx}
\usepackage{multirow}
\usepackage{booktabs}
\usepackage{xcolor}
\usepackage{hyperref}
\usepackage{cite}
\usepackage{algorithm}
\usepackage{algpseudocode}
\usepackage{pifont}
\newcommand{\cmark}{\ding{51}} 
\newcommand{\xmark}{\ding{55}} 

\usepackage[defaultcolor=black]{changes}

\title{Learn by Reasoning: Analogical Weight Generation for Few-Shot Class-Incremental Learning}
\author{Jizhou~Han\textsuperscript{\rm *},
Chenhao~Ding\textsuperscript{\rm *},
Yuhang~He,
Songlin~Dong,
Qiang~Wang,\\
Xinyuan~Gao,
Yihong Gong\textsuperscript{\dag}, \IEEEmembership{Fellow,~IEEE}
\\

\thanks{Jizhou Han, Yuhang He, Qiang Wang and Yihong Gong are with the State Key Laboratory of Human-Machine Hybrid Augmented Intelligence, Institute of Artificial Intelligence and Robotics, Xi'an Jiaotong University, Xi'an 710049, Shaanxi, China; Songlin Dong is with the Faculty of Microelectronics, Shenzhen University of Advanced Technology; Chenhao Ding and Xinyuan Gao are with the College of Software Engineering, Xi'an Jiaotong University, Xi'an 710049, Shaanxi, China.}

\thanks{$^\dagger$ Yihong Gong is the corresponding author. \textsuperscript{\rm *} Jizhou Han and Chenhao Ding are co-first authors.}%

\thanks{ This article has been accepted for publication in IEEE Transactions on Circuits and Systems for Video Technology. This is the author's version which has not been fully edited and 
content may change prior to final publication. Citation information: DOI 10.1109/TCSVT.2025.3637903}%
}

\IEEEpubid{\begin{minipage}{\textwidth}\ \\[30pt] \centering
Copyright \copyright~2025 IEEE. Personal use of this material is permitted. However, permission to use this material for any other purposes must be obtained from the IEEE by sending an email to pubs-permissions@ieee.org.
\end{minipage}}

\begin{document}

\maketitle

\begin{abstract}
    Few-shot class-incremental Learning (FSCIL) enables models to learn new classes from limited data while retaining performance on previously learned classes. Traditional FSCIL methods often require fine-tuning parameters with limited new class data and suffer from a separation between learning new classes and utilizing old knowledge. Inspired by the analogical learning mechanisms of the human brain, we propose a novel analogical generative method. Our approach includes the Brain-Inspired Analogical Generator (BiAG), which derives new class weights from existing classes without parameter fine-tuning during incremental stages. BiAG consists of three components: Weight Self-Attention Module (WSA), Weight \& Prototype Analogical Attention Module (WPAA), and Semantic Conversion Module (SCM). SCM uses Neural Collapse theory for semantic conversion, WSA supplements new class weights, and WPAA computes analogies to generate new class weights. Experiments on miniImageNet, CUB-200, and CIFAR-100 datasets demonstrate that our method achieves higher final and average accuracy compared to SOTA methods.
\end{abstract}


\begin{figure}[t]
    \centering
    \includegraphics[width=0.99\linewidth]{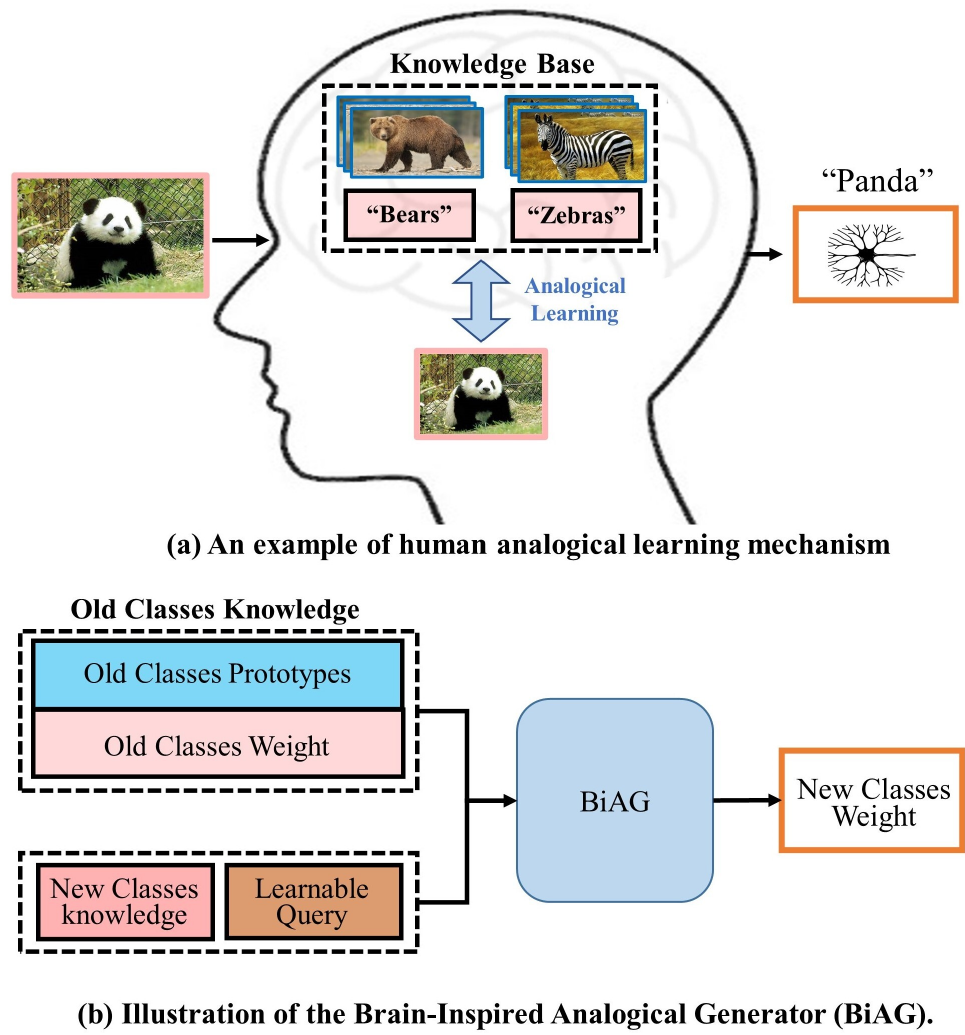}
    \caption{An example of human analogical learning mechanism and the proposed Brain-Inspired Analogical Generator (BiAG). (a) Humans can efficiently form concepts of new classes through analogical reasoning. For example, when encountering a new animal like a panda, one may relate its visual characteristics to previously known classes—such as the body shape of a bear and the black-and-white color pattern of a zebra. This analogy enables the construction of a concept of "panda" even from limited samples. (b) Inspired by this cognitive mechanism, our BiAG module analogically generates classification weights for new classes by leveraging prototypes and weights from old classes along with a learnable query derived from new class knowledge. This allows efficient integration of new classes in few-shot incremental learning.}
    \label{fig:intro}
    \vspace{-0.5cm}
\end{figure}

\section{Introduction}
{\IEEEPARstart{C}{{onventional}} deep learning models have achieved remarkable success in various computer vision and pattern recognition tasks. However, these models are typically developed under a closed-world assumption, where all classes are predefined, and ample labeled data is available for each class. In practice, this assumption rarely holds. New classes may appear continuously over time, while labeled data is often expensive or difficult to obtain. To address this limitation, Class-Incremental Learning (CIL) has been proposed to enable models to learn new classes sequentially while retaining prior knowledge. Although effective in fully supervised settings, most existing CIL methods assume sufficient data for each new class and often require storing exemplars from past sessions. These assumptions greatly limit their applicability in dynamic and data-constrained environments.

In many real-world scenarios, such as identifying rare species, adapting to evolving user behavior, or updating clinical diagnosis systems, new classes often emerge under tight time constraints and with very limited data. In such settings, retraining models from scratch or storing all historical data is often infeasible. Moreover, traditional CIL methods tend to overfit to the few new samples or suffer from forgetting previously learned knowledge. To address these challenges, Few-Shot Class-Incremental Learning (FSCIL) has been proposed~\cite{2020TOPIC}. FSCIL requires models to learn new classes from only a handful of labeled examples while maintaining performance on old classes. Typically, the model is trained in a base session with sufficient data, followed by multiple incremental sessions with only a few examples per new class. Data from previous sessions becomes unavailable, and the model must perform classification over all seen classes. This setting emphasizes the need for methods that can effectively balance stability and plasticity under strict data and time constraints.

FSCIL faces two key challenges: overfitting due to limited new-class samples and catastrophic forgetting of old classes \cite{2020TOPIC}. Existing methods fall into two classes: unified learning and sequential learning. Unified methods optimize for all classes jointly using meta-learning~\cite{chi2022metafscil}, replay~\cite{2023BiDist}, or dynamic networks~\cite{2021CEC} to retain prior knowledge. In contrast, sequential approaches decouple new and old class learning, often focusing on efficient representation\cite{cheraghian2021semantic},~\cite{2022FACT},~\cite{2023SVAClearning}. 
Despite their strengths, both approaches often require fine-tuning, leading to overfitting on limited new data. Moreover, by treating new class learning and old knowledge retention as separate processes, they introduce a disconnect that hinders the balanced integration of past knowledge and new information.

In contrast, humans excel in learning new concepts by leveraging prior knowledge through \emph{analogical reasoning}. Rather than isolating new and old information, humans draw connections between existing knowledge and new information to efficiently construct new concepts~\cite{hofstadter2001analogy,gentner2017analogy}. Cognitive science research highlights that analogical reasoning plays a vital role in human learning. The hippocampus converts short-term memories into long-term memories stored in the cerebral cortex, forming a comprehensive knowledge base~\cite{barron2013online}. When humans encounter new concepts, this knowledge base is accessed to retrieve relevant prior experiences. Regions such as the prefrontal cortex and anterior temporal lobes facilitate the process by drawing analogies between new information and existing knowledge{\cite{bunge2005analogical}}. This analogical reasoning process is further supported by the formation of new synapses, which create neural pathways enabling the brain to integrate and generate new concepts \cite{lamprecht2004structural}. For example, as shown in Fig.~\ref{fig:intro}(a), when learning a new concept like a "panda," humans draw upon prior knowledge of related concepts such as "bear" and "zebra," leveraging similarities to efficiently construct the concept of "panda" even with minimal samples.

Inspired by the analogical learning mechanism of the human brain, we propose a novel analogical generation method that derives new class weights from existing ones. Our method sufficiently utilizes prior knowledge during incremental sessions while eliminating the need for parameter training. The framework includes a feature extractor and a Brain-Inspired Analogical Generator (BiAG). The BiAG takes the prototypes (the centric or mean feature of an identical class) of new and old classes, as well as the classification weights of old classes, as inputs to the module, and outputs the generated classification weights of the new classes. 
The BiAG comprises three core components: the Semantic Conversion Module (SCM), the Weight Self-Attention Module (WSA), and the Weight \& Prototype Analogical Attention Module (WPAA). The SCM facilitates smooth transitions between prototypes and classification weights based on Neural Collapse theory. The WSA enhances knowledge utilization by supplementing new class weights, while the WPAA draws analogies between old and new classes, reorganizing old weights to generate accurate new class weights. Motivationally, WSA can be seen as a preparatory step that refines the new-class representation, emphasizing its salient attributes and aligning it with prior knowledge. This mirrors human cognition, where one first identifies the key features of a new concept before drawing analogies to known classes.
Building upon this foundation, our framework is structured into two distinct phases: the \emph{base session training} and \emph{incremental session inference} phases. During the base session, the feature extractor is trained on the base classes to obtain initial classification weights and prototypes, thereby constructing the knowledge base for analogical learning. Subsequently, the BiAG is trained to cultivate the ability to generate weights for novel classes through simulated incremental learning, wherein we randomly divide the base classes into old and new classes and train the model to generate weights for new classes. In the incremental inference phase, as illustrated in Fig.~\ref{fig:intro}(b), the BiAG generates new class weights by leveraging the prototypes of the new classes alongside the prototypes and weights of the old classes.

Compared to previous methods, the proposed analogical generation approach offers a more efficient and balanced integration of new class learning and old knowledge retention, eliminating the need for parameter fine-tuning during incremental learning stages. The main contributions are summarized as follows:
\begin{itemize}
    \item 
     We propose a novel analogical learning framework for FSCIL inspired by the human brain's analogical reasoning mechanism. By mimicking the cognitive process of drawing analogies between prior knowledge and new information, our framework offers an efficient and adaptive approach to incremental learning.
    \item 
    We develop the BiAG, which uses old class knowledge, new class knowledge, and a learnable query to generate new class weights without parameter fine-tuning during incremental sessions.
    \item 
    Experiments on three benchmark datasets demonstrate that our method achieves higher final and average accuracy compared to the current SOTA methods.
\end{itemize}

\section{Related Work}
\subsection{Class-Incremental Learning (CIL)}
Class-incremental learning (CIL) enables models to continually recognize new classes while preserving knowledge of previously learned ones. The main challenge lies in \textit{catastrophic forgetting}, where learning new tasks interferes with old knowledge. To address this, a variety of strategies have been developed.
Regularization-based methods \cite{li2017learning},~\cite{zenke2017continual},~\cite{kirkpatrick2017overcoming} penalize changes to important parameters to preserve past knowledge. For example, EWC \cite{kirkpatrick2017overcoming} leverages Fisher Information to identify critical weights. More recent efforts introduce forgetting-aware directions and equilibrium constraints to improve stability \cite{wen2025class}.
Replay-based methods \cite{rebuffi2017icarl},~\cite{lopez2017gradient},~\cite{rolnick2019experience},~\cite{chaudhry2019tiny} retain exemplars or use generative models \cite{shin2017continual} to rehearse previous classes. To overcome memory and privacy concerns, recent approaches replace exemplars with mixed features and auxiliary classes \cite{song2024nonexemplar} or use semantic mapping and background calibration to improve alignment \cite{xian2024semantic}. Statistical sampling has also been proposed to simulate past data distributions without real samples \cite{cheng2024efficient}.
Parameter isolation strategies \cite{rusu2016progressive},~\cite{mallya2018packnet},~\cite{serra2018overcoming},~\cite{gao2024beyond} assign different parameter subsets to different tasks, avoiding interference. PackNet \cite{mallya2018packnet}, for instance, uses weight pruning and reallocation to accommodate new tasks efficiently.
Dynamic architecture methods \cite{yoon2018lifelong},~\cite{xu2018reinforced},~\cite{dong2024ceat},~\cite{aljundi2017expert} incrementally grow the network to handle new tasks, though often at the cost of increased complexity. Recent improvements include lightweight incremental frameworks \cite{tao2024lightweighted} and memory-boosted transformers \cite{ni2024moboo}.
In addition, several works enhance representation quality to improve learning efficiency. For example, feature expansion and robust training techniques have proven effective in exemplar-free settings \cite{luo2024representation}, while others leverage external semantic knowledge to guide feature learning and improve generalization \cite{wang2023semantic}. 
{CdCIL~\cite{csvt_hu2022curiosity_CIL} proposes a curiosity-driven sample selection strategy, where uncertainty and novelty metrics are jointly used to identify informative training samples for continual adaptation.}

Although these methods perform well on standard benchmarks, they typically rely on ample labeled data and access to prior tasks—conditions not met in few-shot class-incremental learning (FSCIL). In such scenarios, regularization may underfit, replay can overfit, and dynamic structures may be excessive. These limitations highlight the need for more efficient and generalizable CIL approaches under limited supervision.

\subsection{Few-Shot Class-Incremental Learning (FSCIL)}
Few-shot class-incremental learning (FSCIL) aims to learn new classes from few samples over time without forgetting old ones. Early works such as TOPIC \cite{2020TOPIC} and CEC \cite{2021CEC} adopt feature structure preservation or graph-based adaptation. Meta-learning-based methods like MetaFSCIL \cite{chi2022metafscil} and SPPR \cite{zhu2021self} improve adaptability via task sampling and prototype refinement. SPPR{\cite{zhu2021self}} and IDLVQ \cite{chen2020incremental} enhance prototype-based learning through episodic sampling and vector quantization.
To address representation degradation, FACT \cite{2022FACT} pre-allocates virtual prototypes, while Data-free Replay \cite{liu2022few} replays knowledge to balance class learning. LIMIT \cite{zhou2022few} and CLOM \cite{zou2022margin} employ transformers and margin constraints to align semantics and reduce overfitting. GKEAL \cite{2023gkeal} uses Gaussian kernels for analytic learning, and BiDist \cite{2023BiDist} distills knowledge from dual teachers. SVAC \cite{2023SVAClearning} enhances generalization via virtual classes.
Dynamic model-based approaches include SoftNet \cite{2023subnet}, WaRP \cite{2023warp}, and KRRM \cite{KRRM}, each tailoring model structures for incremental adaptation. NC-FSCIL \cite{2023NC-FSCIL} pre-assigns classifiers based on Neural Collapse, while ALICE \cite{2022Alice} and OrCo \cite{ahmed2024orco} improve intra-class compactness and generalization via angular loss or orthogonality. 
{Comp-FSCIL~\cite{Rebuttal_Comp-FSCIL} further exploits compositional knowledge to enhance few-shot adaptation.}
{YourSelf~\cite{yourself} employs a lightweight Feature Rectification module to refine final features using intermediate ones, guided by relation transfer losses and multi-layer ensemble.}
{Recent methods like PCL~\cite{Rebuttal_PCL} and LRT~\cite{Rebuttal_LRT} use pretrained CLIP models for feature extraction, but risk leaking novel class information via shared embeddings.}
While ALFSCIL \cite{li2024analogical} explores analogical learning, it only leverages prototypes and still relies on fine-tuning. In contrast, our method, BiAG, fully utilizes both prototypes and old class weights to analogically generate new classifiers in a scalable and tuning-free manner.

\subsection{Analogical Learning Mechanism}
Analogical learning is a fundamental aspect of human cognition, allowing individuals to transfer knowledge from familiar domains to novel situations by recognizing relational similarities. This mechanism supports efficient and flexible learning and has been identified by cognitive science as a key driver of human intelligence~\cite{hofstadter2001analogy, gentner2017analogy, gentner2010analogical}. According to Gentner's Structure-Mapping Theory~\cite{gentner1983structure}, analogical reasoning involves retrieving relevant experiences, aligning relational structures (mapping), projecting inferences, evaluating consistency, and re-representing knowledge when necessary.

Neurologically, analogical reasoning is supported by a distributed network involving the hippocampus, prefrontal cortex (PFC), and anterior temporal lobes. The hippocampus transforms short-term experiences into long-term memories stored in the cortex~\cite{barron2013online}, serving as a knowledge base for analogical retrieval. The left frontopolar cortex and inferior PFC aid in semantic processing and relational integration, while the right dorsolateral PFC supports inference and decision-making~\cite{bunge2005analogical}. {At the cellular level, learning relies on structural synaptic plasticity, where the formation and stabilization of new synaptic connections and dendritic spine modifications contribute to the consolidation of long-term memory~\cite{lamprecht2004structural}.} Inspired by these principles, we propose a brain-inspired analogical generator that emulates this mechanism for few-shot class-incremental learning.

\begin{figure*}[t]
    \centering
    \includegraphics[ width=0.95\textwidth]{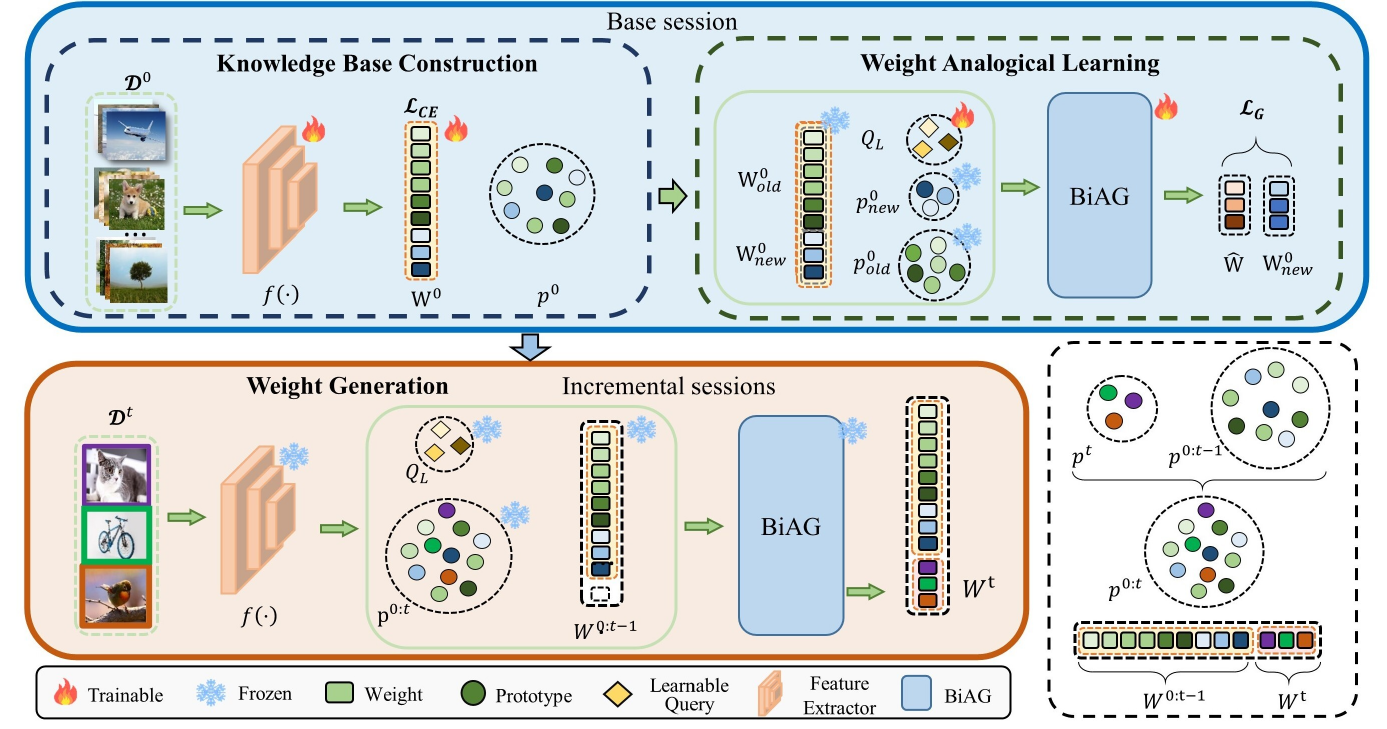}
    \caption{\textbf{An overview of the proposed Analogical Generative Method.} \textit{Knowledge Base Construction}: The network is trained to obtain classification weights and prototypes for the base classes. \textit{Analogical Generator Training}: The BiAG uses pseudo-incremental training and ${\mathcal{L}}_{G}$ to ensure that generated weights match true weights. \textit{Weight Generation}: In the \emph{t}-th session, BiAG generates weights for the new classes using new class prototypes, old class prototypes, and old class weights.}
    \label{fig:main}
\end{figure*}

\section{Method}
\subsection{Problem Setting}
Few-shot class-incremental Learning (FSCIL) aims to continuously learn new classes from a few examples while retaining previously acquired knowledge across multiple sessions. Assume we have a stream of labeled training sets \(\{\mathcal{D}^0, \ldots, \mathcal{D}^t, \ldots, \mathcal{D}^T\}\) and test sets \(\{\mathcal{Z}^0, \ldots, \mathcal{Z}^t, \ldots, \mathcal{Z}^T\}\), where \(T\) is the total number of incremental sessions. The training set \(\mathcal{D}^t = \{(x_i^t, y_i^t) \mid i = 1, \ldots, |\mathcal{D}^t|\}\) contains samples \(x_i^t\) and their corresponding labels \(y_i^t \in \mathcal{C}^t\), where \(\mathcal{C}^t\) represents the classes in the \(t\)-th session. The same applies to the test set \(\mathcal{Z}^t = \{(z_i^t, y_i^t) \mid i = 1, \ldots, |\mathcal{Z}^t|\}\). For all \(i, j\) and \(i \neq j\), \(\mathcal{C}^i \cap \mathcal{C}^j = \varnothing\).
The first training session is called the base session, where the training set \(\mathcal{D}^0\) contains a sufficient number of examples for each class \(c \in \mathcal{C}^0\). For any other session \(t > 0\), \(\mathcal{D}^t\) is an N-way K-shot training set, containing \(N\) classes and \(K\) training examples for each class \(c \in \mathcal{C}^t\). At the end of each training session \(t\), the model is evaluated using the combined test set \(\mathcal{Z}^{0:t} = \mathcal{Z}^0 \cup \cdots \cup \mathcal{Z}^t\).

\subsection{Framework Overview}
As shown in Fig.\ref{fig:main}, our method consists of a feature extractor \(f(\cdot)\), a Brain-Inspired Analogical Generator (BiAG) \(\varphi_{\rm{_{BiAG}}}(\cdot)\), and linear classification weights \(W\).
In the training phase, the \textbf{\textit{Knowledge Base Construction}} and \textbf{\textit{Analogical Generator Training}} stages are involved. Initially, in the \textbf{\textit{Knowledge Base Construction}} stage, the feature extractor \(f(\cdot)\) is first trained on the \(\mathcal{D}^0\) to obtain classification weights and prototypes for the base classes \(\mathcal{C}^0\). In the subsequent \textit{\textbf{Analogical Generator Training}} stage, BiAG is trained to have the capability of generating new class weights through analogy by employing pseudo-incremental training and a loss function \(\mathcal{L}_G\) to ensure that the generated weights closely match the true weights. 
In the incremental phase, corresponding to the\textbf{\textit{ Weight Generation for Newly Emerged Classes}} stage, when new classes emerge, we use the trained feature extractor \(f(\cdot)\) to extract prototypes of the new classes. BiAG \(\varphi_{\rm{_{BiAG}}}(\cdot)\) then generates weights for the new classes by leveraging these new prototypes along with those from old classes and the weights of the old classes. During this phase, the parameters of both BiAG and the feature extractor are frozen.

\subsubsection{Knowledge Base Construction}
This stage simulates the hippocampus's function of converting short-term memories into a long-term repository of relational knowledge. BiAG first constructs a structured knowledge base by training a feature extractor on base classes. This process generates prototypes that encapsulate the central features of each class, providing a foundation for relational mapping. These prototypes act as a semantic bridge, facilitating the integration of new class information without disrupting existing representations.

In the base session, we use $\mathcal{D}^{0}$ that contains sufficient data to train the feature extractor $f\left( \cdot  \right)$ for knowledge base construction. We optimize the model through sample loss during training:
\begin{align}
& \mathcal{L}_{cls} = -\sum_{i=1}^{N_{D^0}} y_i^0 \log\left(\frac{\exp(h_W(f(x_i^0)))}{\sum_{j=1}^{|\mathcal{C}^0|} \exp(h_W(f(x_j^0))) }\right)
\end{align}
\noindent where,  ${\rm{h}}_{W}(\cdot)$ is the linear classification layer and $f\left( \cdot  \right)$ denotes the feature extractor. $N_{D^{0}}$ is the number of the samples of $\mathcal{D}^{0}$. 

After the training is completed, we freeze \(f(\cdot)\) to alleviate catastrophic forgetting and save the classification weights \(W^0\) of base classes for subsequent processes. The feature extractor remains frozen in all subsequent stages.

\begin{algorithm}[t]
  \caption{Our Method Framework}
  \label{alg:biag_framework}
  \begin{algorithmic}[1]
   \State \textbf{Step 1: Knowledge Base Construction (Base Session)}
   \For{each epoch in base session training}
        \State Train feature extractor \(f(\cdot)\) and classifier on \(\mathcal{D}^0\)
   \EndFor
   \State Compute prototypes \(\boldsymbol{p}^0\) using \(f(\cdot)\) \Comment{Eq.(\ref{eq:p})}
   \State Save classification weights \(W^0\) and freeze \(f(\cdot)\)

   \vspace{2mm}
   \State \textbf{Step 2: Analogical Generator Training}
   \For{each epoch in BiAG training}
        \State Randomly split \(\mathcal{C}^0\) into \(\mathcal{C}_{old}^0\) and \(\mathcal{C}_{new}^0\)
        \State Get prototypes \(\boldsymbol{p}_{old}^0\), \(\boldsymbol{p}_{new}^0\), and weights \(W_{old}^0\), \(W_{new}^0\)
        \State Generate weights for pseudo-new classes:
        \[ G \gets \varphi_{\text{BiAG}}(\boldsymbol{p}_{old}^0, \boldsymbol{p}_{new}^0, W_{old}^0)    \] \Comment{Eq.(\ref{eq:3})}
        \State Compute analogical loss:
        \[
        \mathcal{L}_G = 1 - \frac{G \cdot {W_{new}^0}^\top}{\|G\| \cdot \|W_{new}^0\|}  
        \]\Comment{Eq.(\ref{eq:4})}
        \State Backpropagate and update BiAG parameters
   \EndFor
   \State Freeze \(\varphi_{\text{BiAG}}(\cdot)\)
   
\vspace{2mm}
   \State \textbf{Step 3: Weight Generation for New Classes (Sessions \(t=1\) to \(T\))}
   \For{each session \(t=1\) to \(T\)}
        \State Extract prototypes \(\boldsymbol{p}^t\) from \(\mathcal{D}^t\) using frozen \(f(\cdot)\)
        \State Generate new class weights:
        \[
        W^t \gets \varphi_{\text{BiAG}}(\boldsymbol{p}^{0:t-1}, \boldsymbol{p}^t, W^{0:t-1})
        \]\Comment{Eq.(\ref{eq:5})}
        \State Update classifier by concatenating weights:
        \[
        W^{0:t} \gets \text{Concat}(W^{0:t-1}, W^t)
        \]\Comment{Eq.(\ref{eq:6})}
        \State Evaluate on \(\mathcal{Z}^{0:t}\)
   \EndFor
   \State \Return Final classification weights \(W^{0:T}\)
  \end{algorithmic}
\end{algorithm}

\subsubsection{Analogical Generator Training}
In this stage, BiAG is trained to have the capability of generating new class weights through analogy. To achieve this goal, steps are as follows:

First, we obtain prototypes \(\boldsymbol{p}^0\) for all base classes \(\mathcal{C}^0\) using the feature extractor \(f(\cdot)\). The prototypes \(\boldsymbol{p}^0\) are the average embeddings for each class, calculated as follows:
\begin{equation}
{\boldsymbol{p}_{{{c}_{i}}}}=\frac{1}{{{N}_{{{c}_{i}}}}}\sum\nolimits_{j=1}^{{{N}_{{{c}_{i}}}}}{f({{x}_{j,{{c}_{i}}}})}
\label{eq:p}
\end{equation}
\noindent where \(x_{j,c_i}\) is the \(j\)-th sample of class \(c_i\), and \(N_{c_i}\) is the number of samples of class \(c_i\).

Next, in each round of training, we randomly divide the base classes \(\mathcal{C}^0\) into pseudo-old classes \(\mathcal{C}_{old}^0\) and pseudo-new classes \(\mathcal{C}_{new}^0\), ensuring $\mathcal{C}_{old}^0 \cap \mathcal{C}_{new}^0 = \emptyset $. Based on this division, we also split \(\boldsymbol{p}^0\) into \(\boldsymbol{p}^0_{old}\) and \(\boldsymbol{p}^0_{new}\), and \(W^0\) into \(W^0_{old}\) and \(W^0_{new}\).
We define the generated class weights as \(G = \left\{ g_{c_i} \right\}_{i=1}^{|\mathcal{C}_{new}^0|}\) (\(c_i \in \mathcal{C}_{new}^0\)). Then we employ the BiAG to generate the \(G\) with \(\boldsymbol{p}^0_{old}\), \(\boldsymbol{p}^0_{new}\), and \(W^0_{old}\):
\begin{equation}
{\rm{G}}={\varphi}_{\rm{_{BiAG}}}({\boldsymbol{p}^{0}_{old}},{\boldsymbol{p}^{0}_{new}},{{\rm{W}}_{old}^{0}})
\label{eq:3}
\end{equation}
\noindent where \({\varphi}_{\rm{_{BiAG}}}(\cdot)\) denotes BiAG.
To ensure the stability and accuracy of \(G\), we use cosine similarity between \(G\) and \(W_{new}^0\) to constrain \(G\):
\begin{equation}
{\mathcal {L}_{G}}=1-\frac{{\rm{G}} \cdot {{\rm{W}}_{new}^{0}}^{T}}{\|{\rm{G}}\| \cdot \|{\rm{W}}_{{new}}^{0}\|}
\label{eq:4}
\end{equation}
\noindent where \(W_{new}^0\) is the classification weight in \(W^0\) corresponding to class \(\mathcal{C}_{new}^0\).

Additionally, our approach of randomly selecting \(\mathcal{C}_{new}^0\) for multiple rounds of repeated training helps mitigate the bias introduced by the randomness of class selection, thereby enhancing the robustness of analogical generation.

\subsubsection{Weight Generation for New Classes}  
In this stage, both the feature extractor \(f(\cdot)\) and the analogical generator BiAG \({\varphi}_{\rm{_{BiAG}}}(\cdot)\) are frozen to preserve the previously acquired knowledge. The goal is to generate classification weights for new classes using only a few labeled samples. This is achieved by analogically leveraging the semantic and structural knowledge encoded in the prototypes and weights of old classes.

When new classes appear in session \(t\), we first extract the prototypes \(\boldsymbol{p}^t\) of the new classes using the frozen feature extractor \(f(\cdot)\). These prototypes capture the central representation of each new class. At the same time, we maintain the prototypes \(\boldsymbol{p}^{0:t-1} = [\boldsymbol{p}^0, \ldots, \boldsymbol{p}^{t-1}]\) and classification weights \({\rm{W}}^{0:t-1} = \{ {\rm{W}}^i \}_{i=0}^{t-1}\) of all previously seen classes. 

Then, we input the old class knowledge—i.e., \(\boldsymbol{p}^{0:t-1}\) and \({\rm{W}}^{0:t-1}\)—along with the new prototypes \(\boldsymbol{p}^t\) into BiAG to generate the classification weights \({\rm{W}}^t\) for the new classes:
\begin{equation}
{\rm{W}}^t = {\varphi}_{\rm{_{BiAG}}}(\boldsymbol{p}^{0:t-1}, \boldsymbol{p}^t, {\rm{W}}^{0:t-1})
\label{eq:5}
\end{equation}
This process can be viewed as a form of cognitive analogy, where the system understands new classes by linking them to known ones in the learned semantic space.

Compared to fine-tuning or storing exemplars, this strategy provides a flexible and efficient mechanism for incorporating novel classes. It avoids direct parameter updates and minimizes interference with previously learned knowledge, which is crucial for preventing catastrophic forgetting in FSCIL scenarios.

Finally, we update the classifier by concatenating the new weights with the previous weights:
\begin{equation}
{\rm{W}}^{0:t} = {\rm{Concat}}[{\rm{W}}^{0:t-1}, {\rm{W}}^t]
\label{eq:6}
\end{equation}
This updated weight matrix \({\rm{W}}^{0:t}\) is then used for classification across all classes encountered from session 0 to \(t\).

\subsection{Brain-Inspired Analogical Generator}
Inspired by the prefrontal cortex's capacity for abstract reasoning and relational mapping, as shown in Fig.\ref{fig:3}, BiAG integrates three specialized modules to emulate analogical reasoning: the Weight Self-Attention module (WSA), the Weight \& Prototype Analogical Attention (WPAA), and the Semantic Conversion module (SCM). 

\begin{figure}[t]
\centering
\includegraphics[width=0.75\columnwidth]{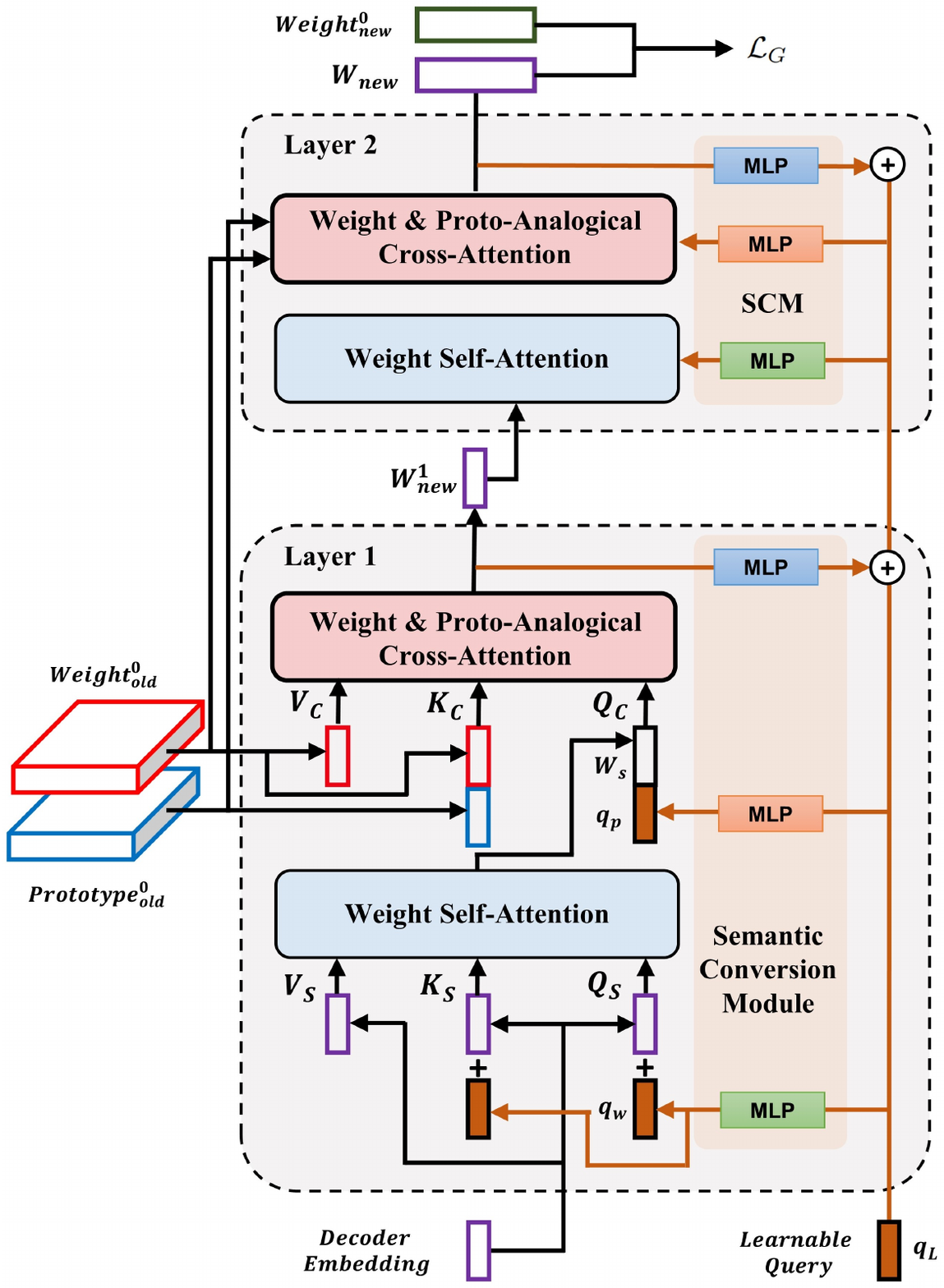} 
\caption{The architecture of the proposed Brain-Inspired Analogical Generator (BiAG). BiAG consists of stacked layers of Weight Self-Attention and Weight \& Prototype Analogical Cross-Attention modules. A shared Semantic Conversion Module (SCM) transforms learnable queries between prototype and weight semantics to facilitate analogical generation of new class weights.}
\label{fig:3}
\end{figure}

\subsubsection{Semantic Conversion Module (SCM)}

The SCM transforms learnable queries between \textit{prototype semantics} and \textit{weight semantics} based on the Neural Collapse (NC) theory~\cite{papyan2020prevalence}. 
According to NC, during the terminal phase of network training, features and classifier weights exhibit the following properties:

\begin{itemize}
    \item {\textbf{NC1 (Feature Collapse):}  
    Let $\mu_{k,i} \in \mathbb{R}^{d}$ denote the feature of the $i$-th sample in class $k$,  
    and let $n_k$ be the number of samples in class $k$.  
    The class mean (prototype) is defined as  
    $\mu_k = \frac{1}{n_k} \sum_i \mu_{k,i}$,  
    and all features of class $k$ collapse to $\mu_k$.}
    \item {\textbf{NC2 (Simplex ETF of Class Means):}  
    Let $\mu_G = \frac{1}{K}\sum_{k=1}^K \mu_k$ denote the global mean over $K$ classes.  
    The centered class means $\widetilde{\mu}_k = \mu_k - \mu_G$  
    form a Simplex Equiangular Tight Frame (ETF),  
    satisfying $\langle \widetilde{\mu}_k, \widetilde{\mu}_{k'} \rangle = -\frac{c^2}{K-1}$ for $k \neq k'$,  
    with $\|\widetilde{\mu}_k\| = c$.}
    \item {\textbf{NC3 (Self-Dual Alignment):}  
    Let $w_k \in \mathbb{R}^d$ denote the classifier weight for class $k$.  
    The weight directions align with the normalized prototypes,  
    $\frac{w_k}{\|w_k\|} = \frac{\widetilde{\mu}_k}{\|\widetilde{\mu}_k\|}$.}
    \item {\textbf{NC4 (Nearest-Mean Prediction):}  
    A sample is classified by the nearest class mean,  
    $\hat{y} = \arg\max_k \langle w_k, \mu \rangle = \arg\min_k \|\mu - \mu_k\|^2$.}
\end{itemize}

These properties indicate that class prototypes $\{\mu_k\}$ and classifier weights $\{w_k\}$  
share the same Simplex-ETF geometry.  
Therefore, there exists an affine mapping between prototypes and weights:  
\begin{equation}
    w_k = s\, {\bf R} \, \widetilde{\mu}_k = {\bf A}\mu_k + {\bf b},
\end{equation}
where $s>0$ is a scalar, ${\bf R}\in \mathbb{R}^{d \times d}$ is an orthogonal matrix,  
${\bf A} = s{\bf R}$, and ${\bf b} = - s{\bf R} \mu_G$.  
This affine relation preserves the NC2–NC3 Simplex-ETF geometry up to a common rotation and scale.
The SCM is implemented as a multi-layer perceptron (MLP) that can exactly represent this affine mapping, thereby preserving the NC2–NC3 geometry between prototypes and weights. Within BiAG, SCM fulfills two key roles: it first converts the learnable query into \textit{weight semantics} for the WSA module to generate supplementary knowledge, and then converts the intermediate outputs into \textit{prototype semantics} for the WPAA, ensuring semantic consistency and enabling effective analogical weight generation.

\subsubsection{Weight Self-Attention (WSA)}
The WSA module is designed to enhance the weights of new classes by supplementing relevant knowledge. WSA selectively focuses on important features in the input data through a self-attention mechanism, thereby improving the representational capacity of the new class weights. Additionally, WSA introduces a learnable query \(\boldsymbol{q}_{L} \in \mathbb{R}^{|\mathcal{C}^t| \times D}\) to further enhance the efficiency of utilizing new class knowledge. The learnable query is initialized with the prototype of the new class \(\boldsymbol{p}^t \in \mathbb{R}^{|\mathcal{C}^t| \times D}\) and transformed into weights by the SCM to ensure the correct semantic representation and accurate expression of new class information:
\begin{equation}
    \boldsymbol{q}_{W} = {\varphi}_{\rm{_{{SCM}}}}(\boldsymbol{q}_{L})
\end{equation}
Next, the WSA module takes the decoder embedding \(\boldsymbol{d}_{E}\) and the obtained \(\boldsymbol{q}_{W}\) as inputs to obtain the supplementary knowledge \(\boldsymbol{W}_{s}\) for the new class weights:
\begin{equation}
  Q_s = K_s =
  \begin{cases}
  q_W + d_E, & n=1,\\
  q_W + \rm{W}^\textit{t}_{\textit{n-1}}, & n>1,
  \end{cases}
\end{equation}

\begin{equation}
  V_s =
  \begin{cases}
  d_E, & n=1,\\
  \rm{W}^\textit{t}_{\textit{n-1}}, & n>1,
  \end{cases}
\end{equation}

\begin{equation}
   {\rm{W}_{s}}={\rm{softmax}} \Big(\frac{{{Q}_{s}}\cdot K_{s}^{T}}{\sqrt{D}}\Big)\cdot {{V}_{s}}. 
\end{equation}
where \(D\) is the dimension of the decoder embedding and \(n\) denotes the layer index in the stacked BiAG block (with \(n{=}1\) being the first layer). WSA enhances the efficiency of knowledge utilization by performing self-attention on \(\boldsymbol{d}_{E}\) and \(\boldsymbol{q}_{W}\), thereby supplementing knowledge related to new class weights. Here, $d_E$ denotes the decoder embedding in BiAG. It is trained only during the \textit{Analogical Generator Training} stage to encode old-class knowledge as a fixed, session-invariant prior and to enable forward-only incremental updates; in all incremental sessions $d_E$ is frozen. At the first layer, $d_E$ is initialized to $\mathbf{0}$ before training to avoid interference from other information; for deeper layers, $d_E$ is taken as the output of the previous layer.

In this design, both $Q_s$ and $K_s$ incorporate the SCM-transformed query $q_W$, ensuring that attention is explicitly guided by new-class semantics, while the value path $V_s$ remains free of $q_L$ to preserve a stable memory content across layers. This separation allows each layer to refine the representation progressively without contaminating the inter-layer content flow.

WSA therefore serves as a preparatory stage in our analogical reasoning framework: it refines the new-class representation by emphasizing salient attributes and aligning them with prior knowledge, akin to how humans first identify the key features of a new concept before drawing analogies to known categories. By augmenting prototypes with supplementary weight-level knowledge, WSA enables new classes to be efficiently learned from minimal data and yields a more complete and semantically expressive representation, thereby providing a stronger foundation for subsequent analogical reasoning in WPAA.

\begin{table*}[t!]
    \caption{Performance of FSCIL in each session on miniImageNet. The results were implemented by \cite{2020TOPIC,2021CEC} in the FSCIL setting. “Average Acc.” is the average accuracy of all sessions. “Average Improv.” indicates the overall improvement of BiAG over each method's average accuracy. “Final Improv.” calculates the improvement of our method in the last session.}
    \small
    \centering
    \setlength{\tabcolsep}{5.2pt}
    \begin{tabular}{lcccccccccccc}
        \toprule
        \multirow{2}{*}{Method} & \multicolumn{9}{c}{Acc. in each session (\%) $\uparrow$} & {Average} & {Average} & {Final} \\
        \cmidrule(lr){2-10}
        ~ & 0 & 1 & 2 & 3 & 4 & 5 & 6 & 7 & 8 & ACC. & Improv. & Improv. \\
        \midrule
        TOPIC (\textit{CVPR}-2020) & 61.31 & 50.09 & 45.17 & 41.16 & 37.48 & 35.52 & 32.19 & 29.46 & 24.42 & 39.64 & \textbf{+30.41} & \textbf{+35.41} \\
        SPPR (\textit{NeurIPS}-2021) & 61.45 & 63.80 & 59.53 & 55.53 & 52.50 & 49.60 & 46.69 & 43.79 & 41.92 & 52.76 & \textbf{+17.29} & \textbf{+17.91} \\
        ERL (\textit{AAAI}-2021) & 61.70 & 56.40 & 54.45 & 51.20 & 47.80 & 45.19 & 44.26 & 41.51 & 39.69 & 49.13 & \textbf{+20.92} & \textbf{+20.14} \\
        CEC (\textit{CVPR}-2021) & 72.00 & 66.83 & 62.97 & 59.43 & 56.70 & 53.73 & 51.19 & 49.24 & 47.63 & 57.75 & \textbf{+12.30} & \textbf{+12.20} \\
        ALICE (\textit{ECCV}-2022) & 80.60 & 70.60 & 67.40 & 64.50 & 62.50 & 60.00 & 57.80 & 56.80 & 55.70 & 63.99 & \textbf{+6.06} & \textbf{+4.13} \\
        CLOM (\textit{NeurIPS}-2022) & 73.08 & 68.09 & 64.16 & 60.41 & 57.41 & 54.29 & 51.54 & 49.37 & 48.00 & 58.48 & \textbf{+11.57} & \textbf{+11.83} \\
        MetaFSCIL (\textit{CVPR}-2022) & 72.04 & 67.94 & 63.77 & 60.29 & 57.58 & 55.16 & 52.90 & 50.79 & 49.19 & 58.85 & \textbf{+11.20} & \textbf{+10.64} \\
        LIMIT (\textit{ECCV}-2022) & 72.32 & 68.47 & 64.30 & 60.78 & 57.95 & 55.07 & 52.70 & 50.72 & 49.19 & 59.06 & \textbf{+10.99} & \textbf{+10.64} \\
        FACT (\textit{CVPR}-2022) & 72.56 & 69.63 & 66.38 & 62.77 & 60.60 & 57.33 & 54.34 & 52.16 & 50.49 & 60.69 & \textbf{+9.36} & \textbf{+9.34} \\
        MCNet (\textit{CVPR}-2023) & 72.33 & 67.70 & 63.50 & 60.34 & 57.59 & 54.70 & 52.13 & 50.41 & 49.08 & 58.64 & \textbf{+11.41} & \textbf{+10.75} \\
        SoftNet (\textit{ICLR}-2023) & 77.17 & 70.32 & 66.15 & 62.55 & 59.48 & 56.46 & 53.71 & 51.68 & 50.24 & 60.86 & \textbf{+9.19} & \textbf{+9.59} \\
        WaRP (\textit{CVPR}-2023) & 72.99 & 68.10 & 64.31 & 61.30 & 58.64 & 56.08 & 53.40 & 51.72 & 50.65 & 59.69 & \textbf{+10.36} & \textbf{+9.18} \\
        GKEAL (\textit{CVPR}-2023) & 73.59 & 68.90 & 65.33 & 62.29 & 59.39 & 56.70 & 54.20 & 52.29 & 51.31 & 60.45 & \textbf{+9.60} & \textbf{+8.52} \\
        BiDist (\textit{CVPR}-2023) & 76.65 & 70.42 & 66.29 & 62.77 & 60.75 & 57.24 & 54.79 & 53.65 & 52.22 & 61.64 & \textbf{+8.41} & \textbf{+7.61} \\
        SVAC (\textit{CVPR}-2023) &81.12 &76.14 &72.43 &68.92 &66.48 &62.95 &59.92 &58.39 &57.11 &67.05 & \textbf{+3.00} & \textbf{+2.72} \\
        ALFSCIL (\textit{TCSVT}-2024) & 81.27 & 75.97 & 70.97 & 66.53 & 63.46 & 59.95 & 56.93 & 54.81 & 53.31 & 64.80 & \textbf{+5.25} & \textbf{+6.52} \\
        NC-FSCIL (\textit{ICLR}-2023) &\underline{84.02} &76.80 &72.00 &67.83 &66.35 &64.04 &61.46 &59.54 &58.31 &67.82 & \textbf{+2.23} & \textbf{+1.52} \\
        KRRM (\textit{TCSVT}-2024) & 82.65 & 77.82 & 73.59 & 70.24 & 67.74 & 64.82 & 61.91 & 59.96 & 58.35 & 68.56 & \textbf{+1.49} & \textbf{+1.48} \\
        YourSelf (\textit{ECCV}-2024) & 84.00 & {77.60} & {73.70} & {70.00} & {68.00} & {64.90} & 62.10& {59.80} & {59.00} & {68.80} & \textbf{+1.25} & \textbf{+0.83} \\
        {Comp-FSCIL (\textit{ICML}-2024)} & {82.78} & {\underline{77.82}} & {\underline{73.70}} & {\underline{70.57}} & {\underline{68.26}} & {\underline{65.11}} & {\underline{62.19}} & {\underline{60.12}} & {\underline{59.00}} & {\underline{68.84}} & {\textbf{+1.21}} & {\textbf{+0.83}} \\
        
        \midrule
        \textbf{BiAG (Ours)} & \textbf{84.78} & \textbf{80.14} & \textbf{75.43} & \textbf{71.48} & \textbf{68.76} & \textbf{65.81} & \textbf{62.99} & \textbf{61.20} & \textbf{59.83} & \textbf{70.05} & -- & -- \\
        \bottomrule
    \end{tabular}
    \label{tab:mini}
\end{table*}

\subsubsection{Weight \& Prototype Analogical Attention (WPAA)}
The role of the WPAA, as shown in Fig.\ref{fig:3}, is to analogize between old knowledge and new knowledge, thereby reorganizing the old classification weights and generating the classification weights for new classes.

First, to enhance generative flexibility and assist in the analogy between new and old knowledge, the $\boldsymbol{{q}_{L}}$ is transformed into prototype semantics $\boldsymbol{{q}_{P}}$ through the SCM and then concatenated with the supplementary knowledge of new class weights ${W}_{s}$ obtained from WSA:
\begin{align}
 &{\boldsymbol{{q}_{P}}={\varphi}_{\rm{_{{SCM}}}}(\boldsymbol{{q}_{L}})},\\
 &\boldsymbol{{z}_{t}}=\rm{Concat}[\rm{W}_{s},{\boldsymbol{{q}_{P}}}].  
\end{align}
Next, as shown in Fig.\ref{fig:3}, we use the obtained ${{z}_{t}}$ as the query for WPAA. The classification weights of the already learned classes ${{\rm{W}}^{0:t-1}}=\left\{ {\rm{W}^{i}} \right\}_{i=0}^{t-1}\in {\mathbb{R}^{|{{\mathcal{C}}^{0:t-1}}|\times \rm{D}}}$ are concatenated with their class prototypes ${\boldsymbol{p}^{0:t-1}}\in{\mathbb{R}^{|{{\mathcal{C}}^{0:t-1}}|\times \rm{D}}}$ to serve as the key for WPAA. The classification weights of the already learned classes ${{\rm{W}}^{0:t-1}}$ are used as the value:
\begin{align}
  & {Q}_{c} = \boldsymbol{{z}_{t}} = \rm{Concat}[\rm{W}_{s},{\boldsymbol{{q}_{P}}}], \\
  & {K}_{c} = \rm{Concat}[{{\rm{W}}^{0:t-1}},{\boldsymbol {p}^{0:t-1}]},\\
  & {V}_{c} = {\rm{W}}^{0:t-1},\\
  & \rm{W}^{t}=\rm {softmax} \Big(
  \frac {{{Q}_{c}}\cdot K_{c}^{T}}{\sqrt{D}}\Big)\cdot {\textit{V}}_{c}.
\end{align}
\noindent where \emph{D} is the dimension of the feature embedding. WPAA calculates the analogy between new and old classes using cross-attention, reorganizing old class weights to generate new class weights. The learnable query is crucial in this process, enhancing knowledge utilization efficiency and ensuring the accurate generation of new class weights.

For scalability, we design WPAA to work together with the SCM. 
Specifically, SCM converts the WPAA output from the previous layer from weight semantics into prototype semantics, and then adds it to the learnable query from the $(n\!-\!1)$-th layer to form the query for the current $n$-th layer:
\begin{align}
  \boldsymbol{{q}_{L_n}}={\varphi}_{\rm{_{SCM}}}(\rm{W}^{t}_{n-1}) + \boldsymbol{{q}_{L_(n-1)}} \
\end{align}
Here, \emph{n} denotes the index of the current layer in BiAG, and $W^{t}_{\,n-1}$ is the WPAA output produced in the previous layer 

The parameters of WSA and WPAA are independent across different layers, 
while SCM shares its parameters globally. This design ensures semantic consistency across layers and progressively refines the learnable query, thus enhancing the model’s analogical reasoning ability through iterative layer-by-layer optimization.

\begin{table*}[t!]
    \caption{Performance of FSCIL in each session on {CUB-200} and comparison with other studies.}
    \centering
    \small
    \setlength{\tabcolsep}{3.2pt}
    \begin{tabular}{lcccccccccccccc}
        \toprule
        \multirow{2}{*}{Method} & \multicolumn{11}{c}{Accuracy in each session (\%) $\uparrow$} & {Average} & {Average} & Final \\
        \cmidrule(lr){2-12}
        ~ & 0 & 1 & 2 & 3 & 4 & 5 & 6 & 7 & 8 & 9 & 10 & ACC. & Improv. & Improv. \\
        \midrule
        TOPIC (\textit{CVPR}-2020) &68.68 &62.49 &54.81 &49.99 &45.25 &41.40 &38.35 &35.36 &32.22 &28.31 &26.28 &43.92 & \textbf{+26.79} & \textbf{+37.44} \\
        ERL (\textit{AAAI}-2021) &73.52 &71.09 &66.13 &63.25 &59.89 &59.49 &58.64 &57.72 &56.15 &54.75 &52.28 &61.17 & \textbf{+9.54} & \textbf{+11.44} \\
        IDLVQ (\textit{NeurIPS}-2020) &77.37 &74.72 &70.28 &67.13 &65.34 &63.52 &62.10 &61.54& 59.04& 58.68& 57.81 &65.23 & \textbf{+5.48} & \textbf{+5.91} \\
        SPPR (\textit{NeurIPS}-2021) &68.68 &61.85 &57.43 &52.68 &50.19 &46.88 &44.65 &43.07 &40.17 &39.63 &37.33 &49.32 & \textbf{+21.39} & \textbf{+26.39} \\
        CEC (\textit{CVPR}-2021) &75.85 &71.94 &68.50 &63.50 &62.43 &58.27 &57.73 &55.81 &54.83 &53.52 &52.28 &61.33 & \textbf{+9.38} & \textbf{+11.44} \\
        MgSvF (\textit{NeurIPS}-2021) &72.29 &70.53 &67.00 &64.92 &62.67 &61.89 &59.63 &59.15 &57.73 &55.92 &54.33 &62.37 & \textbf{+8.34} & \textbf{+9.39} \\
        LIMIT (\textit{ECCV}-2022) &76.32 &74.18 &72.68 &69.19 &68.79 &65.64 &63.57 &62.69 &61.47 &60.44 &58.45 &66.67 & \textbf{+4.04} & \textbf{+5.27} \\
        MetaFSCIL (\textit{CVPR}-2022) &75.90 &72.41 &68.78 &64.78 &62.96 &59.99 &58.30 &56.85 &54.78 &53.82 &52.64 &61.93 & \textbf{+8.78} & \textbf{+11.08} \\
        ALICE (\textit{ECCV}-2022) &77.40 &72.70 &70.60 &67.20 &65.90 &63.40 &62.90 &61.90 &60.50 &60.60 &60.10 &65.75 & \textbf{+4.96} & \textbf{+3.62} \\
        FACT (\textit{CVPR}-2022) &75.90 &73.23 &70.84 &66.13 &65.56 &62.15 &61.74 &59.83 &58.41 &57.89 &56.94 &64.42 & \textbf{+6.29} & \textbf{+6.78} \\
        WaRP (\textit{ICLR}-2023) &77.74 &74.15 &70.82 &66.90 &65.01 &62.64 &61.40 &59.86 &57.95 &57.77 &57.01 &64.66 & \textbf{+6.05} & \textbf{+6.71} \\
        SoftNet (\textit{ICLR}-2023) &78.14 &74.61 &71.28 &67.46 &65.12 &62.39 &60.84 &59.17 &57.41 &57.12 &56.64 &64.56 & \textbf{+6.15} & \textbf{+7.08} \\
        GKEAL (\textit{CVPR}-2023) &78.88 &75.62 &72.32 &68.62 &67.23 &64.26 &62.98 &61.89 &60.20 &59.21 &58.67 &66.34 & \textbf{+4.37} & \textbf{+5.05} \\
        BiDist (\textit{CVPR}-2023) &79.12 &75.37 &72.80 &69.05 &67.53 &65.12 &64.00 &63.51 &61.87 &61.47 &60.93 &67.34 & \textbf{+3.37} & \textbf{+2.79} \\
        ALFSCIL (\textit{TCSVT}-2024) &79.79 &76.53 &73.12 &69.02 &67.62 &64.76 &63.45 &62.32 &60.83 &60.21 &59.30 &66.99 & \textbf{+3.72} & \textbf{+4.42} \\
        NC-FSCIL (\textit{ICLR}-2023) &80.45 &75.98 &72.30 &70.28 &68.17 &65.16 &64.43 &63.25 &60.66 &60.01 &59.44 &67.28 & \textbf{+3.43} & \textbf{+4.28} \\
        KRRM (\textit{TCSVT}-2024) &79.46 &76.11 &73.12 &69.31 &67.97 &65.86 &64.50 &63.83 &62.20 &62.00 &60.97 &67.76 & \textbf{+2.95} & \textbf{+2.75} \\
        {Comp-FSCIL (\textit{ICML}-2024)} & {80.94} & {77.51} & {74.34} & {71.00} & {68.77} & {66.41} & {64.85} & {63.92} & {62.12} & {62.10} & {61.17} & {68.47} & {\textbf{+2.24}} & {\textbf{+2.55}} \\
        SVAC (\textit{CVPR}-2023) &81.85 &\underline{77.92} &74.95 &70.21 &\underline{69.96} &\underline{67.02} &\underline{66.16} &65.30 &63.84 &63.15 &62.49 &69.35 & \textbf{+1.36} & \textbf{+1.23} \\
        YourSelf (\textit{ECCV}-2024) &\textbf{83.40} &77.00 &\underline{75.30} &\textbf{72.20} &69.90 &66.80 &66.00 &\underline{65.60} &\underline{64.10} &\textbf{64.50} &\underline{63.60} &\underline{69.85} & \textbf{+0.86} & \textbf{+0.12} \\
        \midrule
        \textbf{BiAG (Ours)} &\underline{82.97} &\textbf{79.75} &\textbf{76.56} &\underline{71.88} &\textbf{70.72} &\textbf{68.30} &\textbf{68.55} &\textbf{66.49} &\textbf{64.63} &\underline{64.25} &\textbf{63.72} &\textbf{70.71} & -- & -- \\
        \bottomrule
    \end{tabular}
    \label{tab:cub}
\end{table*}

\section{Experiments}
\subsection{Datasets}
Following standard FSCIL benchmarks, we evaluate our method on three widely used datasets: CIFAR-100, miniImageNet, and CUB-200, each with distinct characteristics and evaluation protocols.

\noindent \textbf{CIFAR-100} consists of 100 general object classes with relatively low image resolution. In our setting, the base training set $\mathcal{D}^0$ contains 60 labeled classes, which are used for initial supervised training. The remaining 40 classes are divided into incremental sessions. In each session $t>0$, 5 new classes are introduced with only 5 labeled samples per class, following a 5-way 5-shot format.

\noindent \textbf{miniImageNet} also includes 100 image classes and serves as a popular benchmark for few-shot learning. Similar to CIFAR-100, we adopt 60 classes as the base set $\mathcal{D}^0$ and use the remaining 40 classes for incremental learning. Each incremental session adds 5 new classes, with 5 labeled samples per class, maintaining the 5-way 5-shot configuration.

\noindent \textbf{CUB-200} is a fine-grained dataset containing 200 bird species, characterized by high intra-class similarity and small inter-class variance, making it more challenging for FSCIL. For this dataset, the base training set $\mathcal{D}^0$ consists of 100 known classes. The remaining 100 classes are presented incrementally in 10 sessions, each introducing 10 novel classes with 5 labeled samples per class, following a 10-way 5-shot format.

This three-dataset evaluation protocol allows us to assess model performance under both generic and fine-grained FSCIL scenarios and to evaluate its ability to generalize across different levels of visual granularity with limited supervision.

\begin{table*}[t!]
    \caption{Performance of FSCIL in each session on CIFAR-100 and comparison with other studies.}
    \small
    \centering
   \setlength{\tabcolsep}{5pt}
    \begin{tabular}{lcccccccccccc}
        \toprule
        \multirow{2}{*}{Method} & \multicolumn{9}{c}{Acc. in each session (\%) $\uparrow$} & {Average} & {Average} & {Final} \\
        \cmidrule(lr){2-10}
        ~ & 0 & 1 & 2 & 3 & 4 & 5 & 6 & 7 & 8 & ACC. & Improv. & Improv. \\
        \midrule
        TOPIC (\textit{CVPR}-2020) &64.10 &55.88 &47.07 &45.16 &40.11 &36.38 &33.96 &31.55 &29.37 &42.62 & \textbf{+26.31} & \textbf{+28.58} \\
        ERL (\textit{AAAI}-2021) &73.62 &66.79 &63.88 &60.54 &56.98 &53.63 &50.92 &48.73 &46.33 &57.94 & \textbf{+10.99} & \textbf{+11.62} \\
        SPPR (\textit{NeurIPS}-2021) &64.10 &65.86 &61.36 &57.45 &53.69 &50.75 &48.58 &45.66 &43.25 &54.52 & \textbf{+14.41} & \textbf{+14.70} \\
        LIMIT (\textit{ECCV}-2022) &74.30 &70.44 &66.12 &61.24 &58.77 &53.91 &51.36 &49.20 &45.33 & {58.96} & \textbf{+9.97} & \textbf{+12.62} \\
        C-FSCIL (\textit{CVPR}-2022) &77.47 &72.00 &66.54 &61.39 &58.10 &55.12 &52.47 &50.47 &47.11 & {60.07} & \textbf{+8.86} & \textbf{+10.84} \\
        Data-free Replay (\textit{ECCV}-2022) &74.40 &70.20 &66.54 &62.51 &59.71 &56.58 &54.52 &52.39 &50.14 &60.78 & \textbf{+8.15} & \textbf{+7.81} \\
        MetaFSCIL (\textit{CVPR}-2022) &74.50 &70.10 &66.87 &62.77 &59.48 &56.52 &54.32 &52.56 &49.97 &60.79 & \textbf{+8.14} & \textbf{+7.98} \\
        CLOM (\textit{NeurIPS}-2022) &74.20 &69.83 &66.17 &62.39 &59.26 &56.48 &54.36 &52.16 &50.25 &60.56 & \textbf{+8.37} & \textbf{+7.70} \\
        CEC (\textit{CVPR}-2021) &73.07 &68.88 &65.26 &61.19 &58.09 &55.57 &53.22 &51.34 &49.14 &59.53 & \textbf{+9.40} & \textbf{+8.81} \\
        ALICE (\textit{ECCV}-2022) &79.00 &70.50 &67.10 &63.40 &61.20 &59.20 &58.10 &56.30 &54.10 &63.21 & \textbf{+5.72} & \textbf{+3.85} \\
        FACT (\textit{CVPR}-2022) &74.60 &72.09 &67.56 &63.52 &61.38 &58.36 &56.28 &54.24 &52.10 &62.24 & \textbf{+6.69} & \textbf{+5.85} \\
        WaRP (\textit{ICLR}-2023) &80.31 &76.86 &71.87 &67.58 &64.39 &61.34 &59.15 &57.10 &54.74 &65.92 & \textbf{+3.01} & \textbf{+3.21} \\
        SoftNet (\textit{ICLR}-2023) &80.33 &76.23 &72.19 &67.83 &64.64 &61.39 &59.32 &57.37 &54.94 &66.03 & \textbf{+2.90} & \textbf{+3.01} \\
        GKEAL (\textit{CVPR}-2023) &74.01 &70.45 &67.01 &63.08 &60.01 &57.30 &55.50 &53.39 &51.40 &61.35 & \textbf{+7.58} & \textbf{+6.55} \\
        SVAC (\textit{CVPR}-2023) &78.77 &73.31 &69.31 &64.93 &61.70 &59.25 &57.13 &55.19 &53.12 &63.63 & \textbf{+5.30} & \textbf{+4.83} \\
        BiDist (\textit{CVPR}-2023) &79.45 &75.20 &71.34 &67.40 &64.50 &61.05 &58.73 &56.73 &54.31 &65.41 & \textbf{+3.52} & \textbf{+3.64} \\
        ALFSCIL (\textit{TCSVT}-2024) &80.75 &\underline{77.88} &72.94 &68.79 &65.33 &62.15 &60.02 &57.68 &55.17 &66.75 & \textbf{+2.18} & \textbf{+2.78} \\
       KRRM (\textit{TCSVT}-2024) &81.25 &77.23 &73.30 &69.41 &\underline{66.69} &\underline{63.93} &\underline{62.16} &\underline{59.62} &\underline{57.41} &\underline{67.89} & \textbf{+1.04} & \textbf{+0.54} \\
       
       {Comp-FSCIL(\textit{ICML}-2024)} & {80.93} & {76.52} & {72.69} & {68.52} & {65.50} & {62.62} & {60.96} & {59.27} & {56.71} & {67.08} & {\textbf{+1.85}} & {\textbf{+1.24}} \\
       
        YourSelf (\textit{ECCV}-2024) & \underline{82.90} & 76.30 &72.90 &67.80 &65.20 &62.00 &60.70 &58.80 &{56.60} &67.02 & \textbf{+1.91} & \textbf{+1.35} \\
        NC-FSCIL (\textit{ICLR}-2023) &82.52 &76.82 &\underline{73.34} &\underline{69.68} &66.19 &62.85 & 60.96 & 59.02 &56.11 &67.50 & \textbf{+1.43} & \textbf{+1.84} \\
        \midrule
        \textbf{BiAG (Ours)} &\textbf{84.00} &\textbf{78.97} &\textbf{74.73} &\textbf{70.75} &\textbf{67.36} &\textbf{64.21} &\textbf{62.21} &\textbf{60.20} &\textbf{57.95} &\textbf{68.93} & -- & -- \\
        \bottomrule
    \end{tabular}
    \label{tab:cifar}
\end{table*}

\begin{table}[t!]
    \caption{Performance comparison of different methods.}
    \small
    \centering
     \setlength{\tabcolsep}{4pt}
    \begin{tabular}{lccc}
        \toprule
        \multirow{2}{*}{Method} & Average  & Average  & Average \\
                    & Base Acc. &Final ACC. &Avg Acc.\\
        \midrule
        TOPIC (\textit{CVPR}-2020) & 64.70 & 26.69 & 42.06 \\
        ERL (\textit{AAAI}-2021) & 69.61 & 46.10 & 56.08 \\
        SPPR (\textit{NeurIPS}-2021) & 64.74 & 40.83 & 52.20 \\
        LIMIT (\textit{ECCV}-2022) & 74.31 & 50.99 & 61.56 \\
        CEC (\textit{CVPR}-2021) & 73.64 & 49.68 & 59.54 \\
        ALICE (\textit{ECCV}-2022) & 79.00 & 56.63 & 64.32 \\
        FACT (\textit{CVPR}-2022) & 74.35 & 53.18 & 62.45 \\
        WaRP (\textit{ICLR}-2023) & 77.01 & 54.13 & 63.42 \\
        GKEAL (\textit{CVPR}-2023) & 75.49 & 53.79 & 62.71 \\
        BiDist (\textit{CVPR}-2023) & 78.41 & 55.82 & 64.80 \\
        ALFSCIL (\textit{TCSVT}-2024) & 80.60 & 55.93 & 66.18 \\
        SVAC (\textit{CVPR}-2023) & 80.58 & 57.57 & 66.68 \\
        KRRM (\textit{TCSVT}-2024) & 81.12 & 58.91 & 68.10 \\
        NC-FSCIL (\textit{ICLR}-2023) & 82.33 & 57.95 & 67.53 \\
        {OrCo (\textit{CVPR}-2024) } & {79.66} &{-}  & {63.87} \\
        {Comp-FSCIL (\textit{ICML}-2024)} & {81.55} & {58.96} & {68.13} \\
        YourSelf (\textit{ECCV}-2024) & \underline{83.43} & \underline{59.73} & \underline{68.54} \\
        \midrule
        \textbf{BiAG (Ours)} & \textbf{83.92} & \textbf{60.50} & \textbf{69.90} \\
        \midrule
    \end{tabular}
    \label{tab:all}
\end{table}

\subsection{Implementation Details} We adopt ResNet-12 and ResNet-18 \cite{Resnet2016} as feature extraction backbones in our experiments, following common practice in the FSCIL literature \cite{2023NC-FSCIL}. Specifically, ResNet-12 is employed for experiments on miniImageNet due to its widespread use in few-shot learning settings, while ResNet-18 is used for CIFAR-100 and CUB-200, consistent with prior FSCIL benchmarks.

During the base session training, we apply standard data preprocessing and augmentation techniques to improve generalization. These include random scaling, horizontal flipping, and CutMix \cite{yun2019cutmix,2021CEC}, which have been shown to be effective in reducing overfitting, especially in low-data regimes. The models are optimized using stochastic gradient descent (SGD) with a momentum of 0.9 and a weight decay of $5 \times 10^{-4}$, a widely used configuration in previous FSCIL studies.

For training in the base session, we run 200 epochs on both CIFAR-100 and miniImageNet, using an initial learning rate of 0.1 and a batch size of 128. For the CUB-200 dataset, which is more fine-grained and prone to overfitting, we set the initial learning rate to 0.02 and reduced the batch size to 64 to stabilize training. The learning rate is decayed by a factor of 0.1 at 100 and 150 epochs. All models are trained on a single NVIDIA 3090 GPU using PyTorch.

Following the base session, the analogical generator (BiAG) training stage is conducted for another 200 epochs, where the backbone is frozen and only the BiAG modules are optimized. This stage has substantially lower computational cost per epoch due to the fixed feature extractor, yet ensures sufficient adaptation of the semantic-to-weight mapping before entering the incremental sessions.

This consistent and well-established training setup ensures a fair comparison with prior FSCIL methods while allowing us to focus on evaluating the impact of our proposed analogical learning framework.

\subsection{Comparison with State-of-the-Art Methods}
We benchmark our method against a wide range of state-of-the-art (SOTA) FSCIL methods across three standard datasets: miniImageNet, CIFAR-100, and CUB-200. The detailed session-wise results are reported in Tables~\ref{tab:mini}, \ref{tab:cub} and \ref{tab:cifar}, respectively. These experiments follow the FSCIL protocol introduced in \cite{2020TOPIC}, where each incremental session introduces a small number of novel classes with limited labeled samples. 

\noindent \textbf{On miniImageNet}, as presented in Table~\ref{tab:mini}, our method achieves the highest final session accuracy of 59.83\% and the best average accuracy of 70.05\%. Compared to YourSelf and {NC-FSCIL}, two of the strongest recent methods, BiAG achieves improvements of \textbf{0.83\%} and \textbf{1.52\%} in final accuracy and \textbf{1.25\%} and \textbf{2.23\%} in average accuracy, respectively. These consistent gains demonstrate the effectiveness of our analogical reasoning framework in maintaining discriminative power across incremental sessions, especially in the later stages where performance often degrades.

\noindent \textbf{On CUB-200}, a fine-grained dataset with high intra-class similarity, our method achieves a final session accuracy of 63.72\% and an average accuracy of 70.71\%, as detailed in Table~\ref{tab:cub}. Compared to \textit{SVAC}, BiAG shows an improvement of \textbf{1.23\%} in final accuracy and \textbf{1.36\%} in average accuracy. Notably, BiAG also surpasses YourSelf, indicating its capacity to handle fine-grained incremental learning with stronger semantic consistency. These results highlight that our method not only performs well on general datasets but also excels under more challenging fine-grained scenarios.

\noindent \textbf{On CIFAR-100}, as shown in Table~\ref{tab:cifar}, BiAG obtains a final accuracy of 57.95\% and an average accuracy of 68.93\%, outperforming YourSelf by \textbf{1.35\%} in the final session and \textbf{1.91\%} on average. The performance gain over \textit{NC-FSCIL} is also clear, with improvements of \textbf{1.84\%} in final accuracy and \textbf{1.43\%} in average accuracy. This result reflects the robustness of our approach in more challenging settings with lower-resolution images and greater inter-class diversity, where many existing methods suffer from severe forgetting or limited generalization.

\noindent\textbf{On ImageNet-1K.}  
We evaluate BiAG on the large-scale ImageNet-1K dataset using a 50-way, 10-shot incremental setting with 500 randomly sampled base classes. As shown in Table~\ref{tab:imagenet1k_summary}, BiAG surpasses SoftNet and SVAC by 6.04\% and 1.61\% in Average Accuracy, demonstrating strong scalability and analogical reasoning capability in large-scale FSCIL.

\begin{table}[t]
\small
\setlength{\tabcolsep}{6pt}
\centering
\caption{{Performance on ImageNet-1K.}}
\label{tab:imagenet1k_summary}
\begin{tabular}{lccc}
\toprule
{Method} & {Base Acc.} & {Final ACC.} & {Avg Acc.} \\
\midrule
{SoftNet (\textit{ICLR}-2023)} & {36.21} & {21.21} & {27.51} \\
{SVAC (\textit{CVPR}-2023)} & {\underline{37.74}} & {\underline{26.12}} & {\underline{31.94}} \\
\midrule
{\textbf{BiAG (Ours)} }& {\textbf{37.98}} & {\textbf{28.09}} & {\textbf{33.55}} \\
\bottomrule
\end{tabular}
\end{table}

\begin{table*}[t]
    \caption{Comparison of Last Session base classes accuracy, Last Session new classes accuracy, Last Session new classes average accuracy, and average accuracy on CUB-200. "-": Results not reported in OrCo \cite{ahmed2024orco} }
    \setlength{\tabcolsep}{13pt}
    \small
    \centering
        {
       \begin{tabular}{lccccc}
            \toprule
             \multirow{2}{*}{Method} &Final Session& Final Session &Final Session&Final Session& Average \\
             & Acc. & Base Acc.  &New Avg Acc. &New Acc.  & Acc.\\
            \midrule
            CEC (\textit{CVPR}-2021)      &52.12  &70.46 &33.78 &34.23 &61.33    \\
            FACT  (\textit{CVPR}-2022)        &56.94  &73.90 &39.98 &40.50 &  62.42  \\
            LIMIT  (\textit{ECCV}-2022)    &57.41  &73.60 &41.22&41.80 &66.67    \\
            ALFSCIL (\textit{TCSVT}-2024)       & 59.30  & 74.21 & 44.39 & 45.17 & 66.99  \\
            NC-FSCIL (\textit{ICLR}-2023)     &59.44  &76.19 &42.69 &45.83 &67.28 \\
            OrCo (\textit{CVPR}-2024)         & -  & 66.62& - &\underline{49.25} & 62.36  \\
            SVAC (\textit{CVPR}-2023)      &\underline{62.49}  &\underline{77.65} &\underline{47.35} & 47.68 &\underline{69.35} \\
            \midrule
            \textbf{BiAG (Ours)}              &\textbf{63.72}  &\textbf{78.57} &\textbf{48.87}  &\textbf{49.36} & \textbf{70.71} \\
            \bottomrule
        \end{tabular}
        }
    \label{tab:cub-new}
\end{table*}

\begin{table}[t!]
    \caption{Comparison of final session base and new class accuracy on CIFAR-100. "-" indicates results not reported in \cite{2022FACT,zhou2022few}.}
    \small
    \centering
    \setlength{\tabcolsep}{4.5pt}
    {
    \begin{tabular}{lccc}
        \toprule
        \multirow{2}{*}{Method} & Final Session & Final Session & Final Session \\
         & Accuracy & Base Acc. & New Avg Acc. \\
        \midrule
        CEC      & 49.10 & 67.90 & 20.90 \\
        FACT      & 52.10 & -     & -     \\
        LIMIT    & 51.23 & -     & -     \\
        SVAC      & 53.12 & 73.07 & 23.20 \\
        NC-FSCIL  & \underline{56.11} & \underline{73.98} & \underline{29.30} \\
        \midrule
        \textbf{BiAG (Ours)}          & \textbf{57.95} & \textbf{76.88} & \textbf{29.56} \\
        \bottomrule
    \end{tabular}
    }
    \label{tab:cifa-new}
\end{table}

\begin{table}[t!]
    \caption{Comparison of final session base and new class accuracy on miniImageNet. "-" indicates results not reported in \cite{2022FACT,zhou2022few}.}
    \small
    \centering
    \setlength{\tabcolsep}{4.5pt}
    {
    \begin{tabular}{lccc}
        \toprule
        \multirow{2}{*}{Method} & Final Session & Final Session & Final Session \\
         & Accuracy & Base Acc. & New Avg Acc.\\
        \midrule
        CEC      & 47.67 & 67.97 & 27.37 \\
        FACT      & 50.49 & -     & -     \\
        LIMIT    & 49.19 & -     & -     \\
        SVAC     & 57.11 & 74.64 & \underline{30.82} \\
        NC-FSCIL & \underline{58.31} & \underline{76.30} & \textbf{31.33} \\
        \midrule
        \textbf{BiAG (Ours)}          & \textbf{59.83} & \textbf{79.58} & 30.21 \\
        \bottomrule
    \end{tabular}
    }
    \label{tab:mini-new}
\end{table}

\noindent \textbf{Across three benchmarks} (miniImageNet, CUB-200 and CIFAR-100), we report the average base session accuracy, final session accuracy, and overall average accuracy to comprehensively evaluate the performance of different methods under the FSCIL setting. As summarized in Table~\ref{tab:all}, our method BiAG consistently achieves the best results across all three metrics.
In terms of {Average final session accuracy}, which reflects the model's ability to retain base knowledge while integrating new classes over time, BiAG achieves \textbf{60.50\%}. This clearly exceeds YourSelf and Comp-FSCIL, suggesting that our method suffers less from catastrophic forgetting and is more stable throughout incremental sessions. For the {overall average accuracy}, BiAG again achieves the highest score of \textbf{69.90\%}, demonstrating superior performance across all sessions. This result surpasses both YourSelf and {{Comp-FSCIL}}, achieving relative improvements of \textbf{1.36\%} and \textbf{{1.77\%}}, respectively.
These consistent improvements across all metrics highlight the robustness, generalization, and effectiveness of the proposed analogical reasoning framework. BiAG is not only capable of building strong initial representations but also excels in retaining knowledge and adapting to new classes in both general and fine-grained FSCIL scenarios.

\subsection{Performance on Base and Novel Classes}

\begin{table}[t]
    \caption{Performance of base and new classes on three datasets.}
    \small
    \centering
    \setlength{\tabcolsep}{4.5pt}
        {
       \begin{tabular}{lccc}
            \toprule
        \multirow{2}{*}{Method} & Final Session & Final Session & Final Session \\
         & Accuracy & Base Acc.  & New Avg Acc.\\
            \midrule
            CEC      &49.63  &68.78 &  27.35 \\
             SVAC       &57.57  &75.12 & 33.79  \\
             NC-FSCIL     &\underline{57.95}  &\underline{75.49} &\underline{34.44}  \\
            \midrule
            \textbf{BiAG (Ours)}               &\textbf{60.50}  &\textbf{78.34} &\textbf{36.21} \\
            \bottomrule
        \end{tabular}
        }
    \label{tab:all-new}
\end{table}

\begin{table*}[t]
\centering
\caption{Ablation study on CIFAR-100 showing the individual and combined effects of the WPAA, WSA, and {SCM} modules within BiAG. We report the accuracy for each session, the overall average accuracy (Average ACC), the improvement over the baseline (Average Improv), and the accuracy gain in the final session (Final Improv.).}
\label{tab:cifar_ablation}
\small
\setlength{\tabcolsep}{5.8pt}
\begin{tabular}{ccc ccccccccc ccc}
\toprule
\multirow{2}{*}{WPAA} & \multirow{2}{*}{WSA} & \multirow{2}{*}{{SCM}} & \multicolumn{9}{c}{Acc. in each session (\%) $\uparrow$} & Average & Average & Final \\
\cmidrule(lr){4-12}
~ & ~ & ~ & 0 & 1 & 2 & 3 & 4 & 5 & 6 & 7 & 8 & Acc. & Improv. & Improv. \\
\midrule
\xmark & \xmark & \xmark & 84.00 & 67.32 & 63.24 & 61.84 & 59.59 & 58.24 & 56.80 & 55.97 & 54.92 & 62.44 & -- & -- \\
\cmark & \xmark & \xmark & 84.00 & 70.32 & 64.47 & 65.25 & 62.94 & 61.82 & 60.60 & 58.98 & 57.60 & 65.11 & \textbf{+2.67} & \textbf{+2.68} \\
\cmark & \cmark & \xmark & 84.00 & 73.83 & 71.24 & 68.75 & 64.61 & 63.48 & 61.24 & 60.01 & 57.73 & 67.21 & \textbf{+4.77} & \textbf{+2.81} \\
\cmark & \cmark & \cmark & \textbf{84.00} & \textbf{78.97} & \textbf{74.73} & \textbf{70.75} & \textbf{67.36} & \textbf{64.21} & \textbf{62.21} & \textbf{60.20} & \textbf{57.95} & \textbf{68.93} & \textbf{+6.49} & \textbf{+3.03} \\
\bottomrule
\end{tabular}
\end{table*}

We evaluate BiAG's ability to learn novel classes while preserving old knowledge on three FSCIL benchmarks: miniImageNet, CIFAR-100, and CUB-200. To comprehensively assess performance, we report the following metrics in the final session: 
(1) \textit{Final Session Accuracy}, which reflects the overall accuracy across all classes (base and novel) in the final session;  
(2) \textit{Final Session Base Acc}, the accuracy restricted to base classes, indicating the model's resistance to forgetting;
(3) \textit{Final Session New Avg Acc}, the average accuracy over all novel classes introduced during incremental learning; 
and (4) \textit{Final Session New Acc}, the accuracy on the final 5-way novel classes, reported for CUB-200 to compare with methods like OrCo~\cite{ahmed2024orco} that only provide this metric.

As shown in Table~\ref{tab:all-new}, BiAG achieves the best results across all three datasets, demonstrating superior overall performance and balance. It reaches 60.50\% final session accuracy, 78.34\% base accuracy, and 36.21\% new class average accuracy, outperforming NC-FSCIL by \textbf{2.85\%} on base accuracy and \textbf{1.77\%} on new average accuracy. These results highlight that BiAG provides a robust and scalable framework for FSCIL, exhibiting consistently strong performance in both knowledge retention and novel class generalization. Its effectiveness across diverse datasets and evaluation metrics validates the strength of our analogical learning strategy.

On \textbf{CIFAR-100} (Table~\ref{tab:cifa-new}), BiAG achieves the best performance across all three metrics—final session accuracy, base class accuracy, and novel class average accuracy. It outperforms NC-FSCIL in every aspect, demonstrating a superior balance between knowledge retention and adaptation and validating its robustness.
On \textbf{miniImageNet} (Table~\ref{tab:mini-new}), BiAG also achieves the highest overall and base class accuracy, surpassing NC-FSCIL by \textbf{3.28\%} in base retention. 

On the \textbf{CUB-200} (Table~\ref{tab:cub-new}), BiAG achieves SOTA performance across all evaluation metrics. Compared with SVAC, one of the strongest prior methods, BiAG yields a clear improvement of \textbf{0.92\%} in final session base class accuracy and \textbf{1.52\%} in the average accuracy across all novel classes in the final session. In the final 5-way incremental step, BiAG achieves 49.36\% accuracy, exceeding OrCo, which focuses more on improving performance for the currently added novel classes. These results underline BiAG's effectiveness in fine-grained FSCIL settings, where classes often exhibit high visual similarity. The consistent improvements across both old and new classes suggest that BiAG not only retains previously learned knowledge but also excels at distinguishing subtle differences among closely related new classes.

\subsection{{Comparison with CLIP-based Methods}}

\begin{table}[t]
\centering
\caption{{Performance comparison of our method with recent CLIP-based FSCIL methods on miniImageNet and CIFAR-100. Results are average accuracies across all sessions.}}
\label{tab:pcl_summary_slim}
\small
\setlength{\tabcolsep}{3pt}
\begin{tabular}{lccc}
\toprule
{{Method}} & {{Backbone}} & {{miniImageNet}} & {{CIFAR-100}} \\
\midrule
{CLIP (0-shot)} & {Vis + Text} & {62.67} & {32.24} \\
{LRT (\textit{TPAMI}-2025)} & {Vis + Text} & {{75.94}} & {71.50} \\
{PCL (\textit{TCSVT}-2025)} & {Vis + Text} & {\textbf{94.02}} & \underline{82.92} \\
\midrule
{BiAG (Ours)} & {Vis Only} & \underline{86.43} & {\textbf{84.92}} \\
\bottomrule
\end{tabular}
\end{table}

Table~\ref{tab:pcl_summary_slim} compares our BiAG method with representative CLIP-based methods 
on miniImageNet and CIFAR-100. 
CLIP-based methods (CLIP 0-shot, LRT, and PCL) utilize both visual and text encoders, 
whereas our BiAG relies solely on the visual encoder without any language guidance. 
The reported values correspond to the {average accuracy across all incremental sessions} 
under the standard FSCIL evaluation protocol. 
Despite this, BiAG achieves \textbf{84.92\%} on CIFAR-100, 
surpassing PCL and LRT, and performs competitively on miniImageNet.
This highlights that our analogical weight generation framework provides strong generalization 
even without leveraging large-scale language models.

\subsection{Ablation Study}
To better understand the design and effectiveness of the proposed Brain-Inspired Analogical Generator (BiAG), we conduct extensive ablation studies on three FSCIL benchmark datasets. We focus on evaluating the contributions of the core components, the effect of different network depths, and training configurations. The overall results are illustrated in Fig.~\ref{fig:ablation} and detailed in Table~\ref{tab:cifar_ablation}.

 \begin{figure}[t]
    \centering
    \includegraphics[width=1\linewidth]{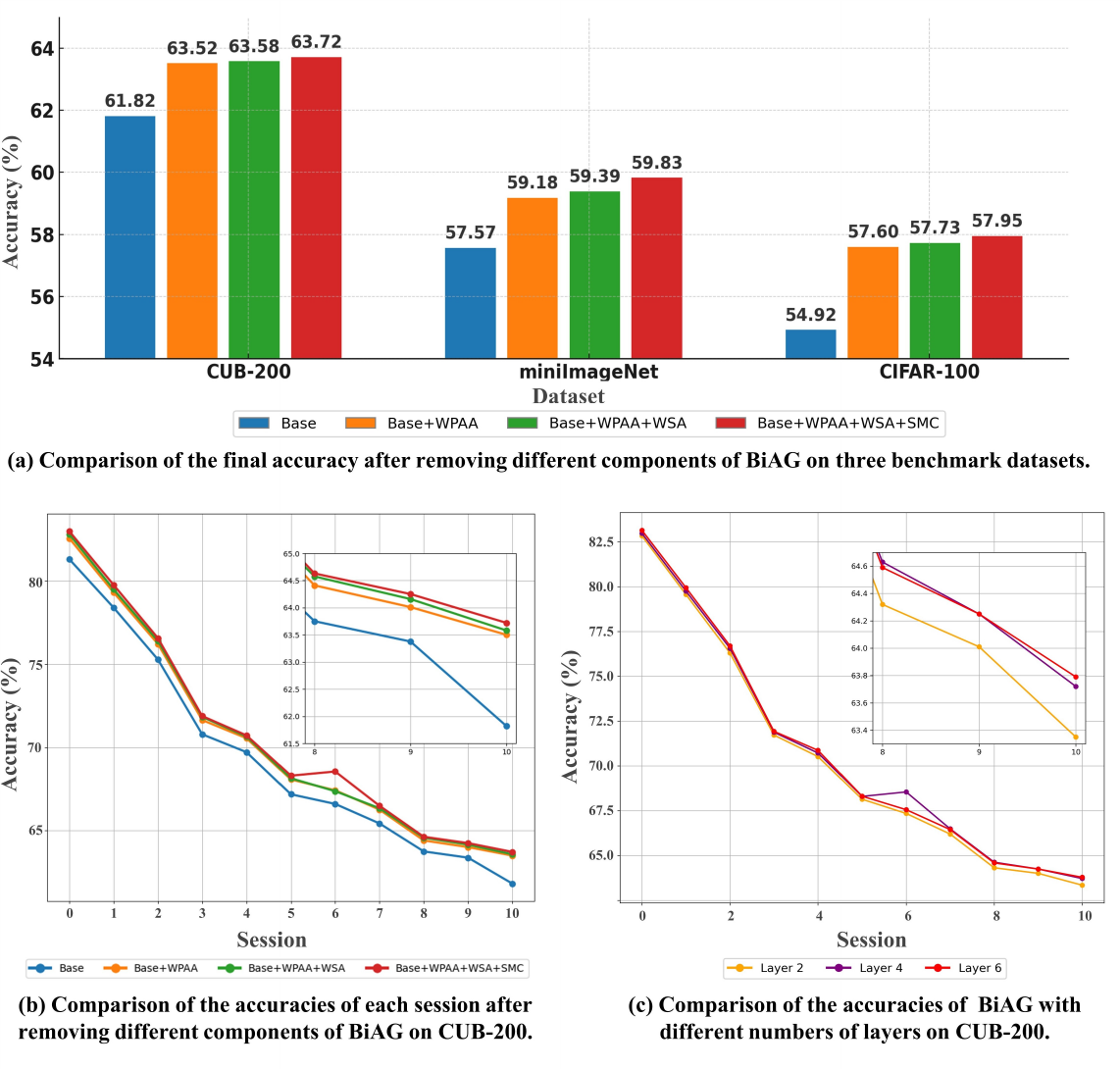}
    \caption{{The results of ablation experiments. Ablation studies on the proposed BiAG.}
    (a) Final session accuracy on three datasets after incrementally adding key components (WPAA, WSA, {SCM}); (b) Session-wise accuracy comparison on CUB-200, showing the contribution of each component throughout the incremental process; (c) Accuracy comparison of BiAG with different numbers of layers, demonstrating that a deeper design improves generalization to new classes.}
    \label{fig:ablation}
\end{figure}

\subsubsection{The Effectiveness of Each Component}
We evaluate the effectiveness of three key modules in BiAG: the Weight \& Prototype Analogical Attention (WPAA), the Weight Self-Attention (WSA), and the Semantic Conversion Module ({SCM}). As illustrated in Fig.~\ref{fig:ablation}(a), the addition of each component leads to a consistent and progressive improvement in final session accuracy across all three benchmarks, confirming their complementary roles.

Among the three, WPAA contributes the most significant performance gain. It provides a mechanism to establish analogical associations between new class prototypes and previously learned semantic structures, enabling effective weight generation even under limited supervision. As shown in Fig.~\ref{fig:ablation}(a), adding WPAA improves the final session accuracy by 1.70\% on CUB-200, 1.61\% on miniImageNet, and 2.68\% on CIFAR-100 over the base model. However, when used in isolation, the improvements diminish in later sessions, suggesting limited temporal stability.
Adding WSA on top of WPAA introduces weight-level relational modeling, which helps the model maintain more consistent predictions as the incremental process progresses. The performance benefits accumulate further with the inclusion of {SCM}, which enhances the semantic compatibility between the feature space and the classifier space. When all three modules are incorporated, BiAG obtains the highest final session accuracy across all benchmarks, achieving cumulative improvements of 1.90\% on CUB-200, 2.26\% on miniImageNet, and 3.03\% on CIFAR-100. On CIFAR-100, replacing SCM’s MLP with a single linear layer within the same WSA$\rightarrow$WPAA stack yields {68.30\%} Average Acc.\ and {57.78\%} Final Acc., which are \textbf{0.63\%} and \textbf{0.17\%} lower than our MLP-based SCM.

The session-wise analysis in Fig.~\ref{fig:ablation}(b) reveals that WPAA significantly improves early-stage performance but suffers from growing degradation without additional modules. WSA alleviates this issue by reinforcing intra-weight consistency across sessions. {SCM} further stabilizes the learning process by aligning high-level semantics between features and weights, especially under growing distributional shifts.
This trend is also supported by the results on CIFAR-100 in Table~\ref{tab:cifar_ablation}. The average accuracy increases from 62.44\% to 65.11\% with WPAA, 67.21\% with WSA, and 68.93\% with all modules enabled—corresponding to an overall improvement of 6.49\%. In the final session, the accuracy grows from 54.92\% to 57.95\%, confirming the effectiveness of our full analogical generation pipeline.

These findings confirm that all three modules play essential and complementary roles. WPAA is fundamental for analogical generation, WSA stabilizes weight interaction over time, and {SCM} improves semantic transfer. Together, they lead to more robust and generalizable weight generation throughout the incremental learning process.

\begin{figure}[t]
    \centering
    \includegraphics[width=0.99\linewidth]{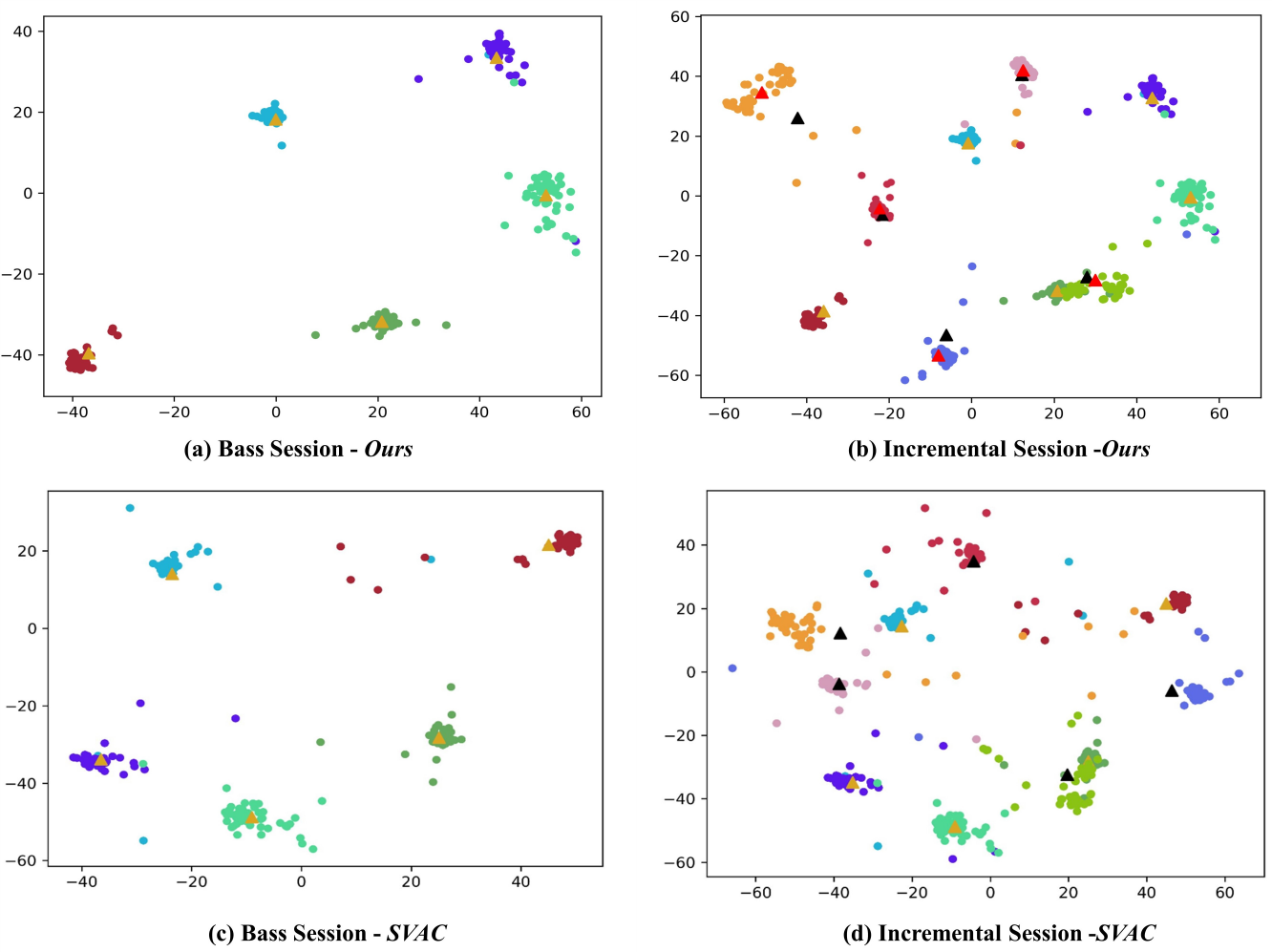}
    \caption{{t-SNE visualization on CIFAR-100 comparing BiAG (Ours) and SVAC. 
    (a) and (c) Base session: clear clusters aligned with classifier weights (yellow). 
    (b) and (d) Incremental session: BiAG-generated weights (red) better align with new class features 
    than prototype-based weights (black), while preserving the structure of base classes.}}
    \label{fig:vis}
\end{figure}

\subsubsection{The Influence of the Number of Layers}
We further investigate the effect of BiAG's model depth by varying the number of stacked layers on the CUB-200 dataset. As shown in Fig.~\ref{fig:ablation}(c), increasing the number of layers leads to consistent improvements in final session accuracy. Specifically, the two-layer variant achieves 63.35\%, while the four-layer and six-layer models reach 63.72\% and 63.79\%. These results suggest that deeper designs help BiAG capture richer analogical patterns and better align new class weights with prior knowledge. However, the performance gain from four to six layers is marginal, indicating diminishing returns as the depth increases. This trend suggests that a four-layer BiAG offers a favorable balance between model capacity and computational efficiency, capturing most of the essential cross-class relational structure without overcomplicating the architecture.

\begin{figure}[t]
    \centering
    \includegraphics[width=0.9\linewidth]{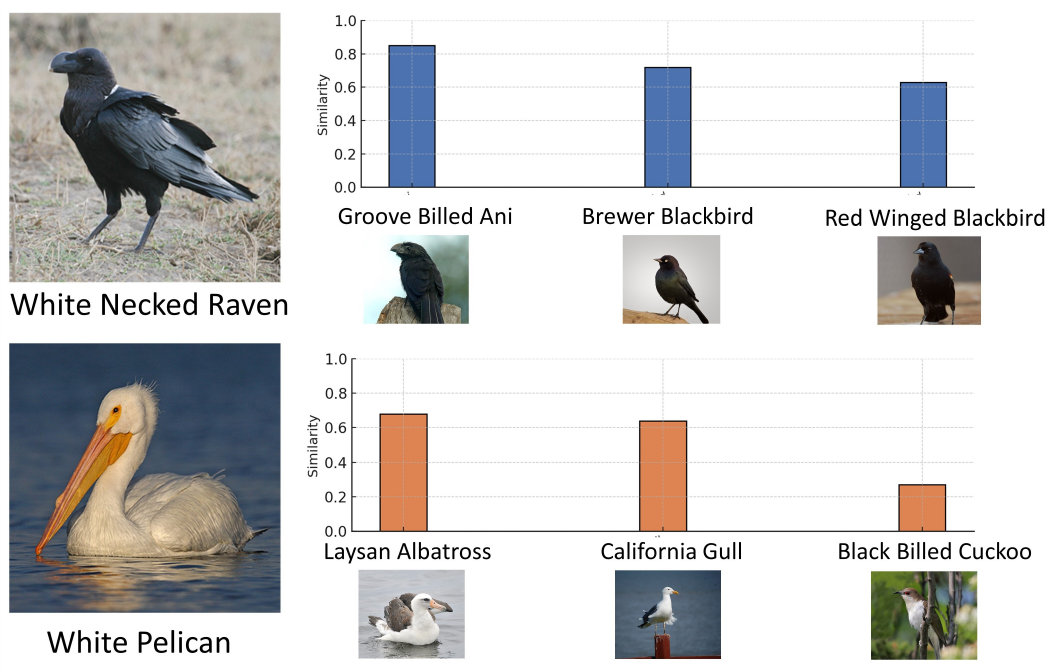}
    \caption{{Visualization of analogical reasoning. We show the base class and the top-3 most similar classes based on the generated weights.}}
    \label{fig:analogy_vis}
\end{figure}

\begin{table*}[t]
\centering
\caption{{Comparison of GPU Memory Usage, Runtime Efficiency, and Average Accuracy Across Different Training Stages on CUB-200}}
\label{tab:complexity_runtime}
\small
\setlength{\tabcolsep}{6.2pt}
\begin{tabular}{lccc|ccc|ccc|c|c}
\toprule
\multirow{2}{*}{{Method}} 
& \multicolumn{3}{c}{{Base Training}} 
& \multicolumn{3}{c}{{BiAG Training}} 
& \multicolumn{3}{c}{{Incremental Training}} 
& \multicolumn{1}{c}{{Total}} 
& \multicolumn{1}{c}{{Avg}} \\
\cmidrule(r){2-4} \cmidrule(r){5-7} \cmidrule(r){8-10}
& {Memory} & {Epochs} & {Time} 
& {Memory} & {Epochs} & {Time} 
& {Memory} & {Epochs} & {Time} 
& {Time} & {Acc} \\
& {(GB)} &        & {(min)} 
& {(GB)} &        & {(min)} 
& {(GB)} & {/Each Session} & {(min)} & {(min)} & {(\%)} \\
\midrule
NC-FSCIL    
& {\textbf{1.45}} & {\textbf{80}} & {56.60} 
& {--}    & {--}   & {--}    
& {\underline{1.45}} & {105--150} & {\underline{3.97}} 
& {60.57} & {67.28} \\

SVAC       
& {18.93} & {\underline{100}} & {\underline{45.36}} 
& {--}    & {--}   & {--}    
& {10.81} & {\underline{10}} & {4.59} 
& {\underline{49.95}} & {69.35} \\

YourSelf    
& {\textbf{-}} & {\textbf{-}} & {-} 
& {--}    & {--}   & {--}    
& 9.51 & {100} &  {212.5}
& {212.5} & {\underline{69.85}} \\

{Ours}        
& {\underline{3.44}} & {200} & {\textbf{23.70}} 
& {\textbf{1.28}} & {\textbf{200}} & {\textbf{10.80}} 
& {\textbf{1.33}} & {\textbf{1}} & {\textbf{2.10}} 
& {\textbf{36.60}} & {\textbf{70.71}} \\
\bottomrule
\end{tabular}
\end{table*}

\subsection{Visualization}

\subsubsection{{Feature Structure}}

Figure~\ref{fig:vis} presents a t-SNE~\cite{van2008visualizing} visualization of feature distributions 
learned on CIFAR-100, comparing BiAG (Ours) and the baseline SVAC. 
These visualizations provide an intuitive understanding of how BiAG organizes the feature space 
across base and incremental sessions and how the generated weights adapt to novel classes 
while retaining base knowledge.

Figures~\ref{fig:vis}(a) and (c) show the base session results. Each point corresponds to a sample, and different colors represent different base classes. Yellow triangles indicate the classifier weights for the base classes. Both methods form clear and well-separated clusters.

Figures~\ref{fig:vis}(b) and (d) illustrate the incremental session results. Black triangles denote weights computed from novel class prototypes, while red triangles represent BiAG-generated weights. Compared to SVAC, BiAG-generated weights align more closely with the corresponding novel class clusters and preserve the base class structure without significant drift. This demonstrates that BiAG can leverage analogical reasoning from old classes to produce accurate and generalizable weights for new classes, which are more representative of the class and less affected by outliers, thus avoiding significant drift.

Overall, these visualizations show that BiAG maintains inter-class separation, reduces the risk of catastrophic forgetting, and enables balanced incremental learning by bridging the semantic gap between base and novel classes in the classifier space.

\begin{figure}[t]
\centering
\includegraphics[width=0.8\linewidth]{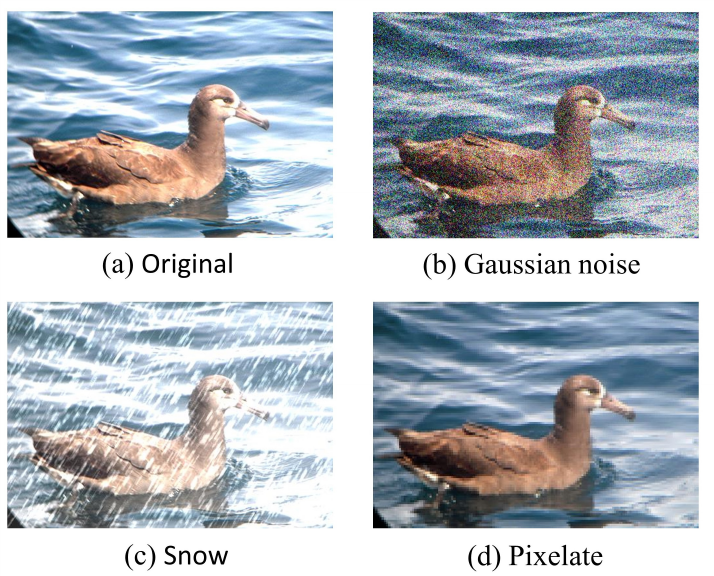}
\caption{{Examples of CUB-200 images under different noise conditions: 
(a) Original clean image, (b) Gaussian noise, (c) Snow, (d) Pixelate.}}
\label{fig:noise_examples}
\end{figure}

\begin{table}[t]
\centering
\caption{{Performance of BiAG on CUB-200 under different noise conditions. 
Final Acc. represents the last session accuracy, and Avg. Acc. is the average across all sessions.}}
\label{tab:noise_test}
\small
\setlength{\tabcolsep}{5pt}
\begin{tabular}{lcc}
\toprule
{{Noise Type}} & {{Final Acc. (\%)}} & {{Avg. Acc. (\%)}} \\
\midrule
{Original }       & {63.72} & {70.71} \\
{Gaussian noise}         & {59.58} & {68.74} \\
{Snow}                   & {55.84} & {62.40} \\
{Pixelate}               & {56.28} & {64.72} \\
\midrule
{\textit{Average (Noisy)}} & {57.23} & {65.29} \\
\bottomrule
\end{tabular}
\end{table}

\begin{table}[t]
\centering
\caption{{Comparison of model parameters and average accuracy on the CUB-200 dataset.}}
\label{tab:param_acc}
\small
\setlength{\tabcolsep}{5pt}
\begin{tabular}{lcc}
\toprule
{{Method}} & {{Params (M)}} & {{Average Acc.}} \\
\midrule
{CEC (CVPR-2021)}           & {12.34} & {61.33} \\
{LIMIT (ECCV-2022)}         & {12.33} & {66.67} \\
{FACT (CVPR-2022)}          & {\textbf{11.34}} & {64.42} \\
{NC-FSCIL (ICLR-2023)}      & {15.91} & {67.28} \\
{ALICE (ECCV-2022)}         & {41.71} & {65.75} \\
{SVAC (CVPR-2023)}          & {24.29} & {69.35} \\
{YourSelf (ECCV-2024)}  & {\underline{11.90}} & {\underline{69.85}} \\
\midrule
{BiAG (Ours)}               & {18.27} & {\textbf{70.71}} \\
\bottomrule
\end{tabular}
\end{table}

\subsubsection{{Analogical Reasoning}}

{To intuitively demonstrate the analogical reasoning capability of our model, we visualize the semantic associations established by our generated weights. Specifically, we select two base classes—\textit{White Necked Raven} and \textit{White Pelican}—which exhibit distinct visual appearances and semantic meanings. For each class, we compute the similarity between its generated weight and all existing base class weights, and visualize the top-3 most similar classes.}

{As shown in Figure~\ref{fig:analogy_vis}, the most similar classes for \textit{White Necked Raven} include \textit{Groove Billed Ani}, \textit{Brewer Blackbird}, and \textit{Red Winged Blackbird}, all of which share dark plumage and similar morphology. In contrast, the generated weight for \textit{White Pelican} finds the closest matches in \textit{Laysan Albatross} and \textit{California Gull}, which share similar aquatic or seabird characteristics.}

{These examples indicate that our model is capable of capturing semantic-level analogies between novel and base classes, even when visual appearances vary significantly. This supports our claim that the generated weights are not arbitrary but stem from a reasoning process grounded in learned concepts.}

\begin{table}[t]
\small
\centering
\caption{{Inference efficiency on CUB-200.}}
\renewcommand{\arraystretch}{1.2}
\setlength{\tabcolsep}{1.6pt}
\begin{tabular}{l cccc}
\hline
{Method} & {Backbone} & {Parameters (M)} & {FLOPs (G)} & {Final Acc.} \\
\hline
{Baseline-R18}      & {ResNet18}  & {\textbf{12.34}} & {\textbf{1.82}} & {61.81} \\
{BiAG (Ours)}   & {ResNet18}  & {\underline{18.27}} & {\underline{2.14}} & {\textbf{63.72}} \\
{SVAC}          & {ResNet18}  & {24.29} & {4.07} & {62.40} \\
{Baseline-R50}      & {ResNet50}  & {26.16} & {4.13} & {\underline{62.56}} \\
{Baseline-R101}      & {ResNet101} & {45.06} & {7.82} & {62.52} \\
\hline
\end{tabular}
\label{tab:cost_comparison}
\end{table}

\subsection{{Robustness under Noisy Conditions}}

To further evaluate the robustness of BiAG, we conducted experiments on CUB-200 under multiple noisy conditions, as illustrated in Figure~\ref{fig:noise_examples} and summarized in Table~\ref{tab:noise_test}. Noisy inputs represent a critical challenge in FSCIL settings because the limited number of novel-class samples makes the model particularly sensitive to input perturbations. BiAG achieves 70.71\% accuracy on clean data and maintains an average of 65.29\% under three different noise types, demonstrating that the proposed analogical weight generation framework preserves strong robustness even when inputs are corrupted.

\subsection{Complexity Analysis }

\subsubsection{{Parameter Comparison}}

To evaluate the efficiency of the proposed Brain-Inspired Analogical Generator (BiAG), we analyze the model complexity in terms of the total number of parameters and its relationship to average accuracy on the CUB-200 dataset. For a fair comparison, all methods are implemented using the ResNet-18 backbone.

Table~\ref{tab:param_acc} presents the comparison of model parameters and the corresponding average accuracy for several representative FSCIL methods. BiAG has 18.27M parameters, which is a moderate level of complexity compared with other approaches.
{Specifically, methods such as FACT, CEC, and LIMIT have 11.34M, 12.34M, and 12.33M parameters, respectively. While these models are lighter, they achieve lower average accuracy than BiAG. Conversely, SVAC and ALICE have 24.29M and 41.71M parameters, significantly larger than BiAG, yet offer comparable or lower accuracy. We also include the recent YourSelf baseline, which uses 11.90M parameters and attains 69.85\% average accuracy on CUB-200. Compared to YourSelf, BiAG achieves +0.86\% higher average accuracy. Our method demonstrates that moderate complexity is sufficient to achieve  performance, balancing representational power and efficiency.

This analysis highlights that BiAG maintains high accuracy without incurring excessive parameter overhead, confirming its practicality for few-shot class-incremental learning scenarios. Future work will explore lightweight extensions of BiAG to further reduce complexity while preserving performance.

\subsubsection{{Training Efficiency}}

We evaluate the runtime and memory efficiency of BiAG on the CUB-200 dataset, as summarized in Table~\ref{tab:complexity_runtime}. BiAG completes the entire training process in 36.6 minutes, which is shorter than SVAC at 49.95 minutes and NC-FSCIL at 60.57 minutes, while requiring only 1.33 GB of memory in the incremental sessions.
For YourSelf, since the authors release trained Base-session weights, its Base Training time is not applicable and is marked as ``--'' in Table~\ref{tab:complexity_runtime}. During incremental sessions, YourSelf requires 212.5 minutes and 9.51 GB of GPU memory, whereas BiAG needs only 2.10 minutes and 1.33 GB, respectively. Consequently, BiAG's total time is 82.8\% shorter than YourSelf.

Although an additional BiAG training stage is introduced, its incremental sessions require only 2.10 minutes each, enabling both efficient training and superior accuracy of 70.71\%. These results indicate that BiAG achieves a favorable balance between computational cost and recognition performance, supporting its practicality for FSCIL tasks. All methods are profiled on a single NVIDIA RTX 3090 GPU to ensure fair timing and memory measurements.

\subsubsection{{Inference Efficiency and Analogical Learning}}

To evaluate whether BiAG’s performance improvements are primarily attributed to its analogical learning mechanism rather than the additional parameters introduced by WSA and WPAA, we analyze parameter efficiency and inference performance across different backbone configurations. The results on the CUB-200 dataset are summarized in Table~\ref{tab:cost_comparison}.

Our BiAG with ResNet-18 achieves 63.72\% final accuracy with only 18.27M parameters and 2.14G FLOPs, outperforming larger backbones such as ResNet-50 (62.56\%) and ResNet-101 (62.52\%), as well as the SOTA method SVAC (62.40\%). This confirms that the performance gain comes primarily from the proposed analogical weight generation mechanism, which effectively leverages semantic relationships between base and novel classes, rather than relying on simply increasing model size or inference cost.

\section{Conclusion and Future Work}
{We proposed the Brain-Inspired Analogical Generator (BiAG), a novel framework for Few-Shot Class-Incremental Learning (FSCIL). Inspired by human analogical reasoning, BiAG generates novel class weights by establishing semantic correspondences with learned knowledge, avoiding parameter updates in incremental sessions. Leveraging old class weights, new prototypes, and a learnable query through its WPAA, WSA, and SCM modules, BiAG achieves strong generalization and reduced forgetting. Extensive experiments on CIFAR-100, miniImageNet, and CUB-200 demonstrate  performance across all key metrics.}

Future work will explore extending analogical generation to domain-incremental and open-vocabulary settings, as well as integrating BiAG with advanced backbones or vision-language models like CLIP. We believe this brain-inspired approach offers a promising direction for flexible and cognitively motivated continual learning systems.

\end{document}